\documentclass[10pt,twocolumn,letterpaper]{article}

\usepackage{cvpr}
\usepackage{times}
\usepackage{epsfig}
\usepackage{amsmath,graphicx}
\usepackage{amssymb}
\usepackage{amsfonts}
\usepackage{subfigure}
\usepackage{multirow}
\usepackage{hhline}
\usepackage{comment}
\usepackage{color}
\DeclareGraphicsExtensions{.eps}
\graphicspath{{.}}

\usepackage[pagebackref=true,breaklinks=true,letterpaper=true,colorlinks,bookmarks=false]{hyperref}

\cvprfinalcopy % *** Uncomment this line for the final submission

\def\cvprPaperID{0007} % *** Enter the CVPR Paper ID here

\ifcvprfinal\fi
\begin{document}

%%%%%%%%%%%%%%%%%%%%%%%%%%%%%%%%%%%%%%%%%%%%%%%%%%%%%%%%
\title{Three Dimensional Fluorescence Microscopy \\ Image Synthesis and Segmentation}
%\title{Three Dimensional Convolutional Neural Networks \\ Based Nuclei Segmentation of Fluorescence Microscopy \\ Images with Generative Adversarial Networks}

\author{Chichen Fu\\
%Video and Image Processing Laboratory\\
%School of Electrical and Computer Engineering\\
Purdue University\\
West Lafayette, Indiana\\
\and
Soonam Lee\\
%Video and Image Processing Laboratory\\
%School of Electrical and Computer Engineering\\
Purdue University\\
West Lafayette, Indiana\\
\and
David Joon Ho\\
%Video and Image Processing Laboratory\\
%School of Electrical and Computer Engineering\\
Purdue University\\
West Lafayette, Indiana\\
\and
Shuo Han\\
%Video and Image Processing Laboratory\\
%School of Electrical and Computer Engineering\\
Purdue University\\
West Lafayette, Indiana\\
\and
Paul Salama\\
%Department of Electrical and Computer Engineering\\
Indiana University-Purdue University\\
Indianapolis, Indiana\\
\and
Kenneth W. Dunn\\
%Division of Nephrology\\
%School of Medicine\\
Indiana University\\
Indianapolis, Indiana\\
\and
Edward J. Delp\\
%Video and Image Processing Laboratory\\
%School of Electrical and Computer Engineering\\
Purdue University\\
West Lafayette, Indiana\\
}

\maketitle
\thispagestyle{empty}	% make this page number empty

%%%%%%%%%%%%%%%%%%%%%%%%%%%%%%%%%%%%%%%%%%%%%%%%%%%%%%%%
%\vspace{-0.05in}
\begin{abstract}
Advances in fluorescence microscopy enable acquisition of 3D image volumes with better image quality and deeper penetration into tissue. Segmentation is a required step to characterize and analyze  biological structures in the images and recent 3D segmentation using deep learning has achieved promising results. One issue is that deep learning techniques require a large set of groundtruth data which is impractical to annotate manually for large 3D microscopy volumes. This paper describes a 3D deep learning nuclei segmentation method using synthetic 3D volumes for training. A set of synthetic volumes and the corresponding groundtruth are generated using spatially constrained cycle-consistent adversarial networks. Segmentation results demonstrate that our proposed method is capable of segmenting nuclei successfully for various data sets.
\end{abstract}
% \begin{keywords}
% nuclei segmentation, fluorescence microscopy, 3D convolutional neural network, synthetic data generation, generative adversarial networks
% \end{keywords}

%%%%%%%%%%%%%%%%%%%%%%%%%%%%%%%%%%%%%%%%%%%%%%%%%%%%%%%%
%\vspace{-0.05in}
\section{Introduction}
\vspace{-0.05in}
\label{sec:intro}

Fluorescence microscopy is a type of an optical microscopy that uses fluorescence to image 3D subcellular structures \cite{vonesch2006,dunn2002}. Three dimensional segmentation is needed to quantify and characterize cells, nuclei or other biological structures.

Various nuclei segmentation methods have been investigated in the last few decades. Active contours \cite{kass1988, delgado2015} which minimizes an energy functional to fit desired shapes has been one of the most successful methods in microscopy image analysis. Since active contours uses the image gradient to evolve a contour to the boundary of an object, this method can be sensitive to noise and highly dependent on initial contour placement. In \cite{li2007} an external energy term which convolves a controllable vector field kernel with an image edge map was introduced to address these problems. In \cite{chan2001} 2D region-based active contours  using image intensity to identify a region of interest was described. This achieves better performance on noisy image and is relatively independent of the initial curve placement. Extending this to 3D, \cite{lorenz2013} described 3D segmentation of a rat kidney structure. This technique was further extended to address the problem of 3D intensity inhomogeneity \cite{lee2017}. However, these energy functional based methods cannot distinguish various structures. %\cite{lee2015}.
Alternatively, \cite{paul2013,rizk2014} described a method known as Squassh to solve the energy minimization problem from a generalized linear model to couple image restoration and segmentation. In addition, \cite{srinivasa2009} described multidimensional segmentation using random seeds combined with multi-resolution, multi-scale, and region-growing technique.

\begin{figure*}[htbp!]
\centerline{\epsfig{figure=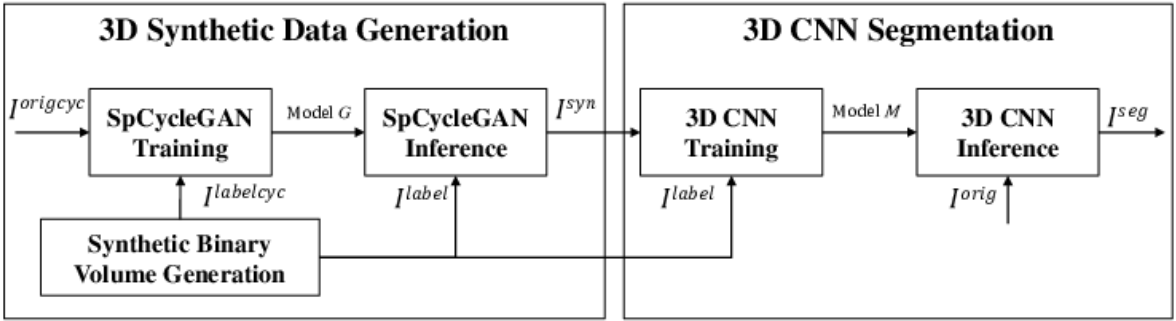,width=0.9\textwidth}}
%\vspace{-0.1in}
\caption{Block diagram of the proposed approach for 3D nuclei segmentation}
\vspace{-0.1in}
\label{fig:diagram}
\end{figure*}

Convolutional neural network (CNN) has been used to address problems in segmentation and object identification \cite{long2015}. Various approaches, based on CNNs, have been used in the biomedical area \cite{litjens2017}. U-Net \cite{ronneberger2015} is a 2D CNN which uses an encoder-decoder architecture with skip connections to segment cells in light microscopy images. In \cite{raza2017} a multi-input multi-output CNN for cell segmentation in fluorescence microscopy images to segment various size and intensity cells was described. Since these approaches \cite{ronneberger2015, raza2017} are 2D segmentation methods, they may fail to produce reasonable segmentation in 3D. More specifically, stacking these 2D segmentation images into 3D volume may result in misalignment in the depth direction \cite{lorenz2013}. 
%A combination of a 2D CNN nuclei segmentation and 3D refinement processing was used for 3D segmentation in \cite{fu2017}. 
Also, in \cite{prasoon2013} a method that trained three networks from different directions in a volume and combined these three results to produce a form of 3D segmentation was described. A 3D U-Net \cite{cciccek2016} was introduced to identify 3D structures by extending the architecture of \cite{ronneberger2015} to 3D. However, this approach requires manually annotated groundtruth to train the network. Generating groundtruth for 3D volumes is tedious and is generally just done on 2D slices, obtaining true 3D groundtruth volumes are impractical. One way to address this is to use synthetic ground truth data \cite{zhang2015, barbosa2017}. A method that segments nuclei by training a 3D CNN with synthetic microscopy volumes was described in \cite{ho2017}. Here, the synthetic microscopy volumes were generated by blurring and noise operations.
%\textbf{{\color{red} You need to re-write the previous sentence. You cannot use this language for anonymous review. The paper will be rejected. Check the entire paper to make sure you do not cite our previous work this way!!}}

Generating realistic synthetic microscopy image volumes remains a challenging problem since various types of noise and biological structures with different shapes are present and need to be modeled. Recently, in \cite{goodfellow2014} a generative adversarial network (GAN) was described to address image-to-image translation problems using two adversarial networks, a generative network and a discriminative network. In particular, the discriminative network learns a loss function to distinguish whether the output image is real or fake whereas the generative network tries to minimize this loss function. 
One of the extensions of GANs is Pix2Pix \cite{isola2017} which uses conditional GANs to learn the relationship between the input image and output image that can generate realistic images. One issue with Pix2Pix \cite{isola2017} is that it still requires paired training data to train the networks. In \cite{liu2016} coupled GANs (CoGAN) for learning the joint distribution of multi-domain images without having the corresponding groundtruth images was introduced. Later, cycle-consistent adversarial networks (CycleGAN) \cite{zhu2017} employed a cycle consistent term in the adversarial loss function for image generation without using paired training data. More recently, a segmentation method using concatenating segmentation network to CycleGAN to learn the style of CT segmentation and MRI segmentation was described in \cite{xu2017}.

In this paper, we present a 3D segmentation method to identify and segment nuclei in fluorescence microscopy volumes without the need of manual segmented groundtruth volumes. Three dimensional synthetic training data is generated using spatially constrained CycleGAN. A 3D CNN network is then trained using 3D synthetic data to segment nuclei structures. Our method is evaluated using hand segmented groundtruth volumes of real fluorescence microscopy data from a rat kidney. Our data are collected using two-photon microscopy with nuclei labeled with Hoechst 33342 staining.

%%%%%%%%%%%%%%%%%%%%%%%%%%%%%%%%%%%%%%%%%%%%%%%%%%%%%%%%
\section{Proposed Method}
\label{sec:method}
\vspace{-0.05in}

Figure \ref{fig:diagram} shows a  block diagram of our method. We denote $I$ as a 3D image volume of size $X \times Y \times Z$. 
Note that $I_{z_p}$ is a $p^\text{th}$ focal plane image, of size  $X \times Y$, along the $z$-direction in a volume, where $p \in \{1, \dots, Z\}$. Note also that $I^{orig}$ and $I^{seg}$ is the original fluorescence microscopy volume and segmented volume, respectively. 
In addition, let $I_{\left(q_i:q_f,r_i:r_f,p_i:p_f\right)}$ be a subvolume of $I$, whose $x$-coordinate is $q_i \leq x \leq q_f$, $y$-coordinate is $r_i \leq y \leq r_f$, $z$-coordinate is $p_i \leq z \leq p_f$, where $q_i, q_f \in \{1, \dots, X\}$, $r_i, r_f \in \{1, \dots, Y\}$, $p_i, p_f \in \{1, \dots, Z\}$, $q_i \leq q_f$, $r_i \leq r_f$, and $p_i \leq p_f$. 
For example, $I^{seg}_{\left(241:272,241:272,131:162\right)}$ is a subvolume of a segmented volume, $I^{seg}$, where the subvolume is cropped between 241$^\text{st}$ slice and 272$^\text{nd}$ slice in $x$-direction, between 241$^\text{st}$ slice and 272$^\text{nd}$ slice in $y$-direction, and between 131$^\text{st}$ slice and 162$^\text{nd}$ slice in $z$-direction.
%\textbf{{\color{red} Please check the notation to make sure this is correct and make sense! Is this the standard notation we developed??!}}
 
As shown in Figure \ref{fig:diagram}, our proposed method consists of two steps: 3D synthetic data generation and 3D CNN segmentation. We first generate synthetic binary volumes, $I^{labelcyc}$, and then use them with a subvolume of the original image volumes, $I^{origcyc}$, to train a spatially constrained CycleGAN (SpCycleGAN) and obtain a generative model denoted as model $G$. This model $G$ is used with another set of synthetic binary volume, $I^{label}$, to generate corresponding synthetic 3D volumes, $I^{syn}$. For 3D CNN segmentation, we can utilize these paired $I^{syn}$ and $I^{label}$ to train a 3D CNN and obtain model $M$. Finally, the 3D CNN model $M$ is used to segment nuclei in $I^{orig}$ to produce $I^{seg}$.

% 3D U-Net Figure
\begin{figure*}[h]
\centerline{\epsfig{figure=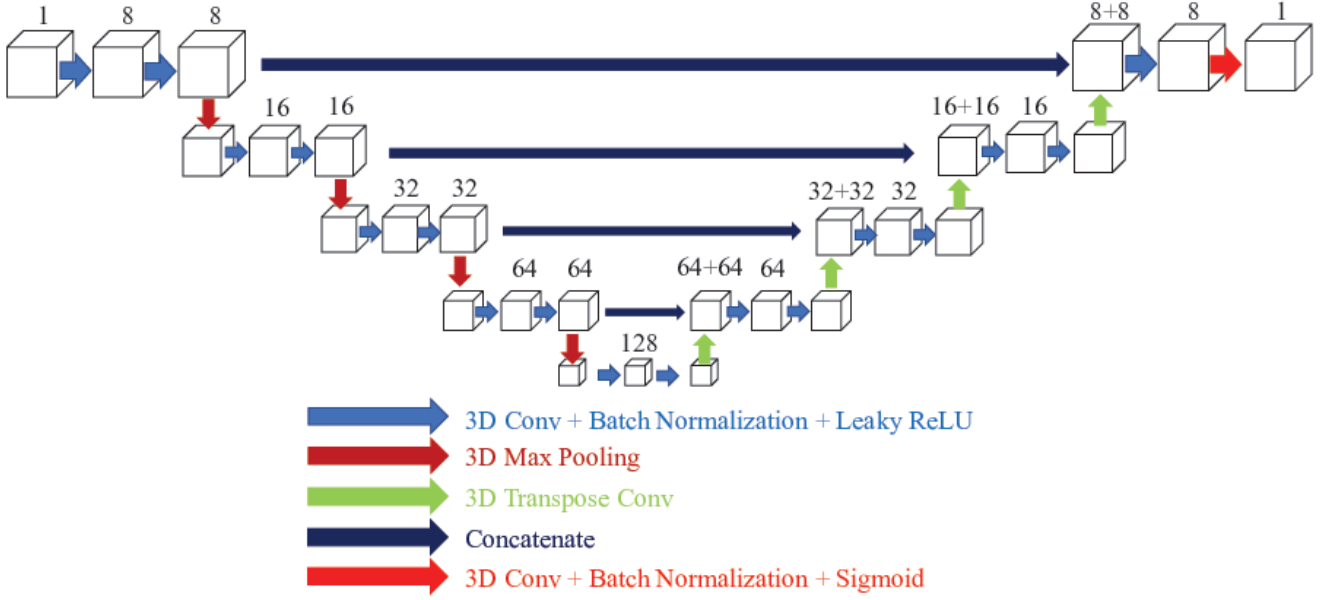,width=1\textwidth}}
\caption{Architecture of our modified 3D U-Net}
\label{fig:3DUNet}
\end{figure*}

\subsection{3D Synthetic Data Generation}
\label{sec:synthetic}
%\vspace{-0.05in}

Three dimensional synthetic data generation consists of synthetic binary volume generation, SpCycleGAN training, and SpCycleGAN inferences. In synthetic binary volume generation, nuclei are assumed to have an ellipsoidal shape, multiple nuclei are randomly generated in different orientations and locations in a volume \cite{ho2017}. The original CycleGAN and our SpCycleGAN were trained to generate a set of synthetic volumes.

\subsubsection{CycleGAN}
\label{sec:originalsynthetic}
The CycleGAN is trained to generate a synthetic microscopy volume. CycleGAN uses a combination of discriminative networks and generative networks to solve a minimax problem by adding cycle consistency loss to the original GAN loss function as \cite{goodfellow2014, zhu2017}:
\vspace{-0.1in}
\begin{align}\label{eq:LCycleGAN}
\mathcal{L}(G, F, D_1, D_2) &= \mathcal{L}_{\text{GAN}}(G,D_1,I^{labelcyc}, I^{origcyc}) \nonumber \\
&+ \mathcal{L}_{\text{GAN}}(F,D_2,I^{origcyc},I^{labelcyc}) \nonumber \\ 
&+ \lambda \mathcal{L}_{cyc}(G,F, I^{origcyc},I^{labelcyc})
\end{align}
where
\\
\resizebox{1.00\hsize}{!}
{ \begin{minipage}{\linewidth}
\begin{align}
\mathcal{L}_{\text{GAN}}(G,D_1,I^{labelcyc},I^{origcyc}) &= \mathbb{E}_{I^{origcyc}}[\text{log}({D_1 (I^{origcyc})})] \nonumber \\
&+ \mathbb{E}_{I^{labelcyc}}[\text{log}(1-{D_1(G(I^{labelcyc})))}] \nonumber
\end{align}
\end{minipage} 
} \\
\resizebox{1.00\hsize}{!}
{ \begin{minipage}{\linewidth}
\begin{align}
\mathcal{L}_{\text{GAN}}(F,D_2,I^{origcyc},I^{labelcyc}) &= \mathbb{E}_{I^{labelcyc}}[\text{log}({D_2 (I^{labelcyc})})] \nonumber \\
&+ \mathbb{E}_{I^{origcyc}}[\text{log}(1-{D_2(F(I^{origcyc})))}] \nonumber  
\end{align}
\end{minipage} 
} \\
\resizebox{1.00\hsize}{!}
{ \begin{minipage}{\linewidth}
\begin{align}
\mathcal{L}_{cyc}(G,F, I^{origcyc},I^{labelcyc}) &= \mathbb{E}_{I^{labelcyc}}[|| F(G(I^{labelcyc})) - I^{labelcyc} ||_1] \nonumber \\
&+ \mathbb{E}_{I^{origcyc}}[|| G(F(I^{origcyc})) - I^{origcyc} ||_1]. \nonumber
\end{align}
\end{minipage} 
} \vspace{0.03in} \\
Here, $\lambda$ is a weight coefficient and $|| \cdot ||_1$ is $L_1$ norm. Note that Model $G$ maps $I^{labelcyc}$ to $I^{origcyc}$ while Model $F$ maps $I^{origcyc}$ to $I^{labelcyc}$. Also, $D_1$ distinguishes between $I^{origcyc}$ and $G(I^{labelcyc})$ while $D_2$ distinguishes between $I^{labelcyc}$ and $F(I^{origcyc})$.
%In particular, our proposed method trains two generative networks, model $G$ and model $F$, and two discriminative networks, $D_1$ and $D_2$. Model $G$ maps $I^{labelcyc}$ to $I^{origcyc}$ while Model $F$ maps $I^{origcyc}$ to $I^{labelcyc}$. Also, $D_1$ distinguishes between $I^{origcyc}$ and $G(I^{labelcyc})$ while $D_2$ distinguishes between $I^{labelcyc}$ and $F(I^{origcyc})$.} 
$G(I^{labelcyc})$ is an original like microscopy volume generated by model $G$ and $F(I^{origcyc})$ is generated by model $F$ that looks similar to a synthetic binary volume. Here, $I^{origcyc}$ and $I^{labelcyc}$ are unpaired set of images. 
%To be more specific, the networks try to minimize the mean square error between $I^{origcyc}$ and $G(I^{labelcyc})$ and maximize the output of the discriminative network which is the probability of $G(I^{labelcyc})$ being identified as $I^{origcyc}$ simultaneously.
In CycleGAN inference, $I^{syn}$ is generated using the model $G$ on $I^{label}$. As previously indicated $I^{syn}$ and $I^{label}$ are a paired set of images. Here, $I^{label}$ is served as a groundtruth volume corresponding to $I^{syn}$.
%\textbf{{\color{red} Again, Please check the notation to make sure this is correct and make sense! Is this the standard notation we developed??!}}

\subsubsection{Spatially Constrained CycleGAN}
Although the CycleGAN uses cycle consistency loss to constrain the similarity of the distribution of $I^{origcyc}$ and $I^{syn}$, CycleGAN does not provide enough spatial constraints on the locations of the nuclei. CycleGAN generates realistic synthetic microscopy images but a spatial shifting on the location of the nuclei in $I^{syn}$ and $I^{label}$ was observed. To create a spatial constraint on the location of the nuclei, a network $H$ is added to the CycleGAN and takes $G(I^{labelcyc})$ as an input to generate a binary mask, $H(G(I^{labelcyc}))$. Here, the architecture of $H$ is the same as the architecture of $G$. Network $H$ minimizes a $L_2$ loss, $\mathcal{L}_{\text{Spatial}}$, between $H(G(I^{labelcyc}))$ and $I^{labelcyc}$. $\mathcal{L}_{\text{Spatial}}$ serves as a spatial regulation term in the total loss function. The network $H$ is trained together with $G$. The loss function of the SpCycleGAN is defined as:
\vspace{-0.1in}
\small
\begin{align}\label{eq:LSpCycleGAN}
\mathcal{L}(G, F, H, D_1, D_2)\! &= \mathcal{L}_{\text{GAN}}(G,D_1,I^{labelcyc}, I^{origcyc}) \nonumber \\
& + \mathcal{L}_{\text{GAN}}(F,D_2,I^{origcyc},I^{labelcyc}) \nonumber \\
& + \lambda_1 \mathcal{L}_{cyc}(G,F,I^{origcyc},I^{labelcyc}) \nonumber \\
& + \lambda_2 \mathcal{L}_{spatial}(G,H,I^{origcyc},I^{labelcyc}) 
\end{align}
\normalsize
where $\lambda_1$ and $\lambda_2$ are the weight coefficients for $\mathcal{L}_{cyc}$ and $\mathcal{L}_{spatial}$, respectively. Note that first three terms are the same and already defined in Equation (\ref{eq:LCycleGAN}). Here, $\mathcal{L}_{spatial}$ can be expressed as

%\resizebox{1.00\hsize}{!}
%{ \begin{minipage}{\linewidth}
%\begin{align}
%&\mathcal{L}(G, F, H, D_1, D_2) = \mathcal{L}_{\text{GAN}}(G,D_1,I^{labelcyc}, I^{origcyc}) + \mathcal{L}_{\text{GAN}}(F,D_2,I^{origcyc},I^{labelcyc}) \nonumber \\
%&+ \lambda_1 \mathcal{L}_{cyc}(G,F, I^{origcyc},I^{labelcyc}) + \lambda_2 \mathcal{L}_{Spatial}(G,I^{origcyc},I^{labelcyc})
%\end{align}
%\end{minipage} 
%}

\resizebox{1.00\hsize}{!}
{ \begin{minipage}{\linewidth}
\begin{align}
\mathcal{L}_{spatial}(G, H, I^{origcyc},I^{labelcyc}) &= \mathbb{E}_{I^{labelcyc}}[|| H(G(I^{labelcyc})) - I^{labelcyc} ||_2] \nonumber.
\end{align}
\end{minipage} 
} \vspace{0.03in} \\

\subsection{3D U-Net}
Figure \ref{fig:3DUNet} shows the architecture of our modified 3D U-Net. The filter size of each 3D convolution is $3 \times 3 \times 3$. To maintain the same size of volume during 3D convolution, a voxel padding of $1 \times 1 \times 1$ is used in each convolution. A 3D batch normalization \cite{ioffe2015} and a leaky rectified-linear unit activation function are employed after each 3D convolution. 
In the downsampling path, a 3D max pooling uses $2 \times 2 \times 2$ with stride of 2 is used. In the upsampling path, feature information is retrieved using 3D transpose convolutions. Our modified 3D U-Net is one layer deeper than conventional U-Net as can be seen in Figure \ref{fig:3DUNet}. Our training loss function can be expressed as a linear combination of the Dice loss ($\mathcal{L}_{Dice}$) and the binary cross-entropy loss ($\mathcal{L}_{BCE}$) such that
\begin{equation}\label{eq:Lseg}
\mathcal{L}_{seg}(T,S) = \mu_1 \mathcal{L}_{Dice}(T,S) + \mu_2 \mathcal{L}_{BCE}(T,S)
\end{equation}
where
\begin{align}
\mathcal{L}_{Dice}(T,S) &= \frac{2 (\sum_{i=1}^{N} t_i s_i)}{\sum_{i=1}^{N} t_i^2 + \sum_{i=1}^{N} s_i^2} \nonumber \\
\mathcal{L}_{BCE}(T,S) &= -\frac{1}{N} \sum_{i=1}^{N} t_i \log(s_i) + (1 - t_i) \log(1-s_i) \nonumber,
\end{align}
respectively \cite{Milletari2016}. Note that $T$ is the set of the targeted groundtruth values and $t_i \in T$ is a targeted groundtruth value at $i^{th}$ voxel location. Similarly, $S$ is a probability map of binary volumetric segmentation and $s_i \in S$ is a probability map at $i^{th}$ voxel location. Lastly, $N$ is the number of entire voxels and $\mu_1$, $\mu_2$ serve as the weight coefficient between to loss terms in Equation (\ref{eq:Lseg}). The network takes a grayscale input volume with size of $64 \times 64 \times 64$ and produces an voxelwise classified 3D volume with the same size of the input volume. To train our model $M$, $V$ pairs of synthetic microscopy volumes, $I^{syn}$, and synthetic binary volumes, $I^{label}$, are used.

\subsubsection{Inference}
For the inference step we first zero-padded $I^{orig}$ by $16$ voxels on the boundaries. A 3D window with size of $64 \times 64 \times 64$ is used to segment nuclei. Since the zero padded $I^{orig}$ is bigger than the 3D window, the 3D windows is slided to $x$, $y$, and $z$-directions by $32$ voxels on zero-padded $I^{orig}$ \cite{ho2017}. Nuclei partially observed on boundaries of the 3D window may not be segmented correctly. Hence, only the central subvolume of the output of the 3D window with size of $32 \times 32 \times 32$ is used to generate the corresponding subvolume of $I^{seg}$ with size of $32 \times 32 \times 32$. 
This process is done until the 3D window maps an entire volume.

%%%%%%%%%%%%%%%%%%%%%%%%%%%%%%%%%%%%%%%%%%%%%%%%%%%%%%%%
\section{Experimental Results}
\label{sec:results}
%\vspace{-0.1in}

We tested our proposed method on two different rat kidney data sets. These data sets contain grayscale images of size $X = 512 \times Y = 512$. Data-I
%\footnote{Data-I was provided by Malgorzata Kamocka of the Indiana Center for Biological Microscopy.} 
consists of $Z = 512$ images, Data-II consist of $Z = 64$. 

Our SpCycleGAN is implemented in Pytorch using the Adam optimizer \cite{kingma2014} with default parameters given by CycleGAN \cite{zhu2017}. In addition, we used $\lambda_1 = \lambda_2 = 10$ in the SpCycleGAN loss function shown in Equation (\ref{eq:LSpCycleGAN}). We trained the CycleGAN and SpCycleGAN to generate synthetic volumes for Data-I and Data-II, respectively. A $128 \times 128 \times 128$ synthetic binary volume for Data-I denoted as $I^{labelcyc_{Data-I}}$ and a $128 \times 128 \times 300$ subvolume of original microscopy volume of Data-I denoted as $I^{origcyc_{Data-I}}$ were used to train model $G^{Data-I}$. 
Similarly, a $128 \times 128 \times 128$ synthetic binary volume for Data-II denoted as $I^{labelcyc_{Data-II}}$ and a $128 \times 128 \times 32$ subvolume of original microscopy volume of Data-II denoted as $I^{origcyc_{Data-II}}$ were used to train model $G^{Data-II}$. 

We generated $200$ sets of $128 \times 128 \times 128$ synthetic binary volumes, $I^{label_{Data-I}}$ and $I^{label_{Data-II}}$ 
where $I^{label_{Data-I}}$ and $I^{label_{Data-II}}$ are generated according to different size of nuclei in Data-I and Data-II, respectively. By using the model $G^{Data-I}$ on $I^{label_{Data-I}}$, $200$ pairs of synthetic binary volumes, $I^{label_{Data-I}}$, and corresponding synthetic microscopy volumes, $I^{syn_{Data-I}}$, of size of $128 \times 128 \times 128$ were obtained. 
Similarly, by using model $G^{Data-II}$ on $I^{label_{Data-II}}$, $200$ pairs of $I^{label_{Data-II}}$ and corresponding $I^{syn_{Data-II}}$, of size of $128 \times 128 \times 128$ were obtained. 
Since our modified 3D U-Net architecture takes volumes of size of $64 \times 64 \times 64$, we divided $I^{label_{Data-I}}$, $I^{syn_{Data-I}}$, $I^{label_{Data-II}}$, and $I^{syn_{Data-II}}$ into adjacent non overlapping $64 \times 64 \times 64$. 
Thus, we have $1600$ pairs of synthetic binary volumes and corresponded synthetic microscopy volumes per each data to train our modified 3D U-Net. Note that these $1600$ synthetic binary volumes per each data are used as groundtruth volumes to be paired with corresponding synthetic microscopy volumes. Model $M^{Data-I}$ and $M^{Data-II}$ are then generated.

Our modified 3D U-Net is implemented in Pytorch using the Adam optimizer \cite{kingma2014} with learning rate $0.001$. For the evaluation purpose, we use different settings of using 3D synthetic data generation methods (CycleGAN or SpCycleGAN), different number of pairs of synthetic training volume $V$ ($V = 80$ or $V = 1600$) among $1600$ pairs of synthetic binary volume corresponding synthetic microscopy volume. Also, we use different loss functions with different settings of the $\mu_1$ and $\mu_2$. Moreover, we also compared our modified 3D U-Net with 3D encoder-decoder architecture \cite{ho2017}. Lastly, small objects which are less than $100$ voxels were removed using 3D connected components.       

\begin{figure}[htb!]
\centering
\subfigure[]
{
   \epsfig{figure=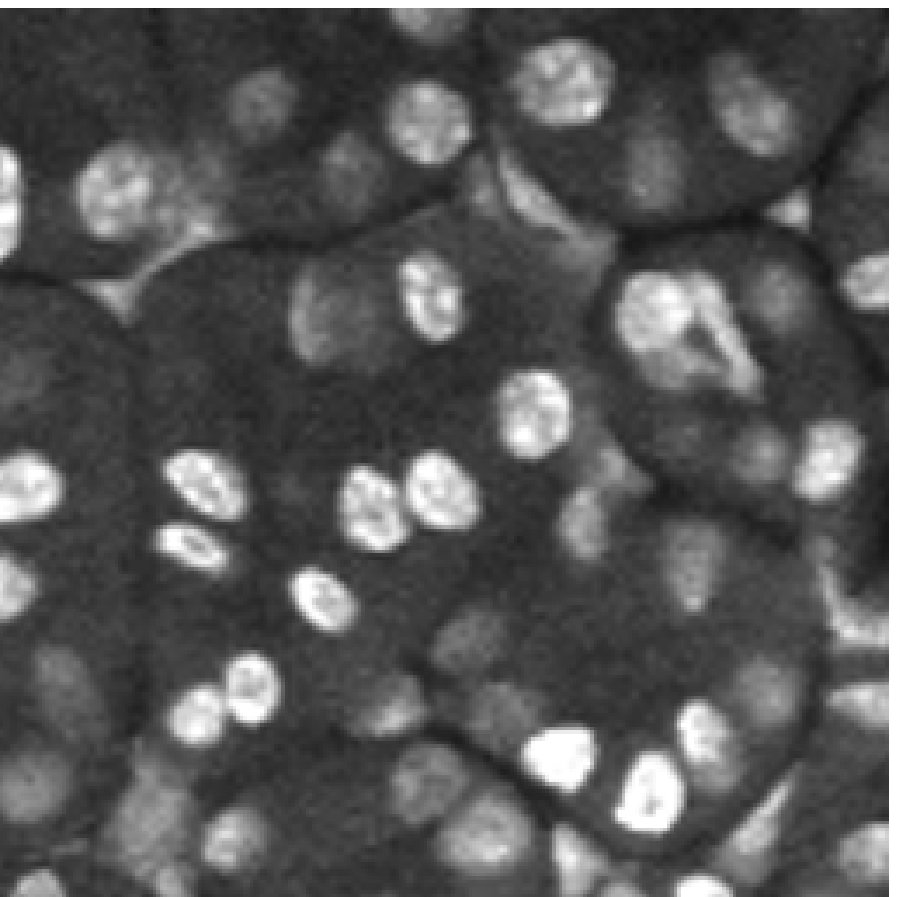,width=2.5cm}
}
\subfigure[]
{
   \epsfig{figure=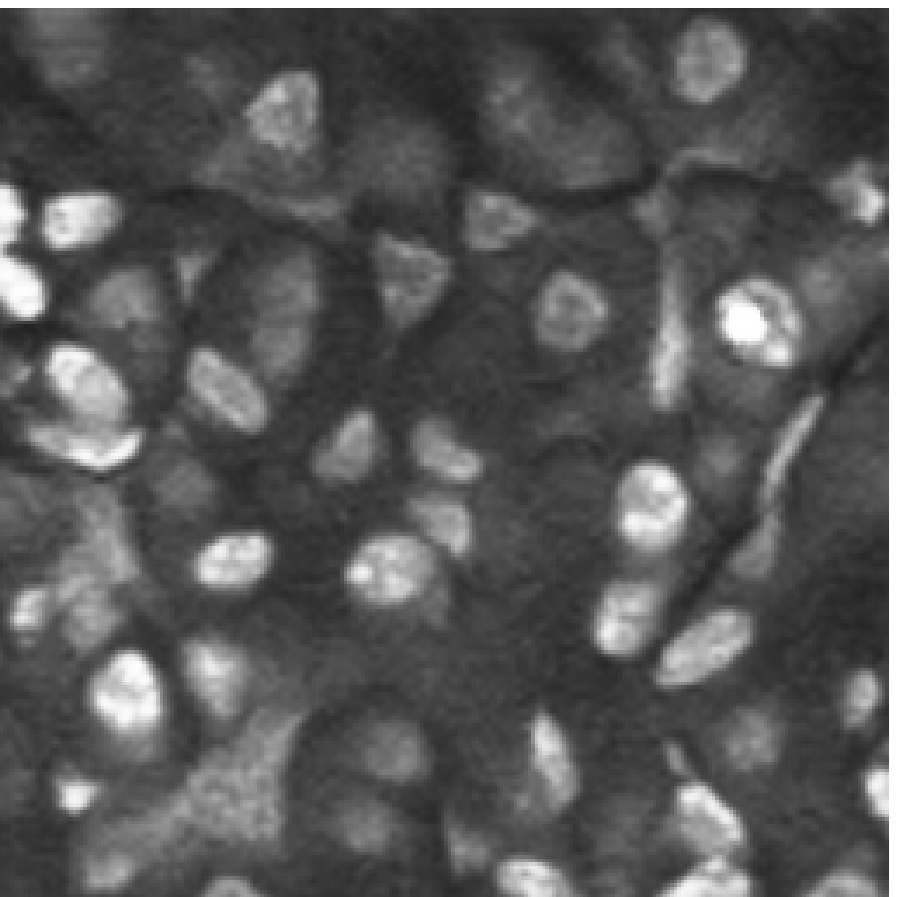,width=2.5cm} %width=3.8cm
}
\subfigure[]
{
   \epsfig{figure=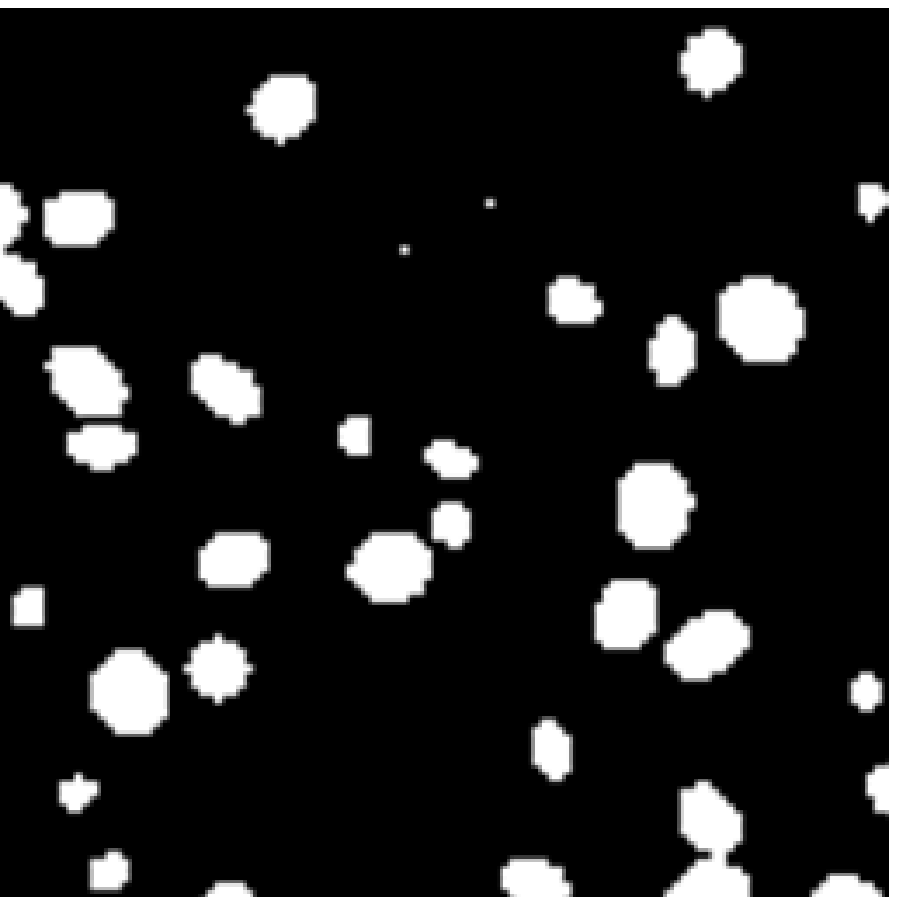,width=2.5cm}
} 
%\vspace{-0.1in}
\subfigure[]
{
   \epsfig{figure=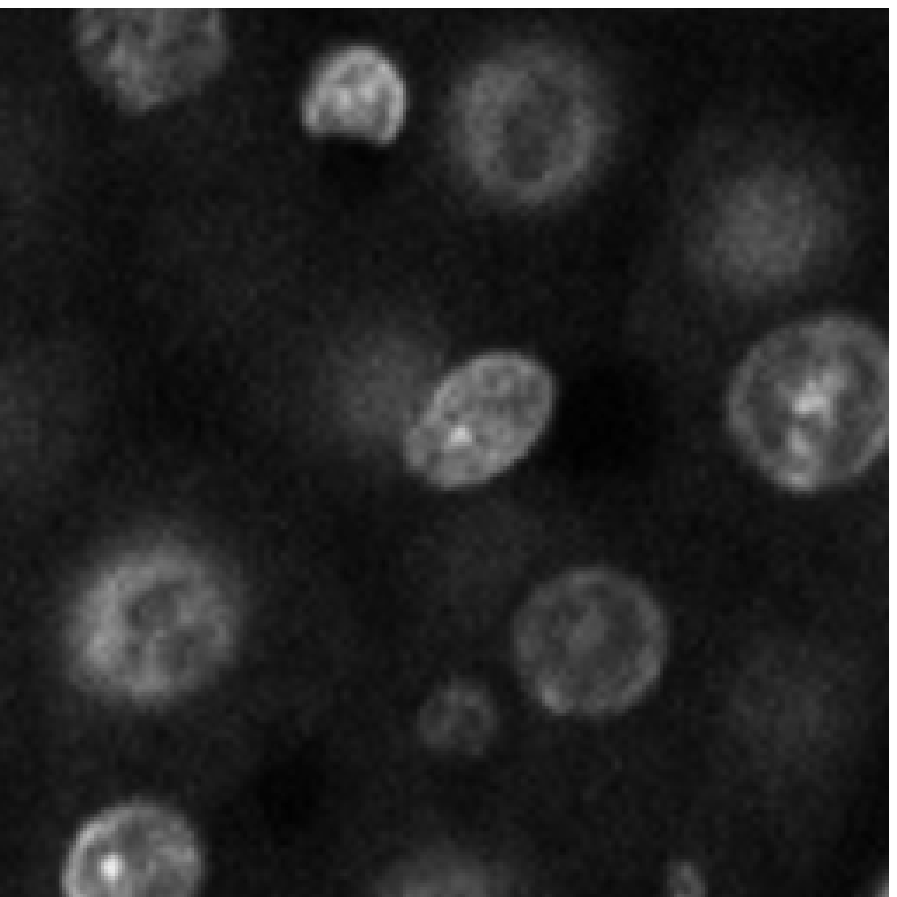,width=2.5cm}
}
\subfigure[]
{
   \epsfig{figure=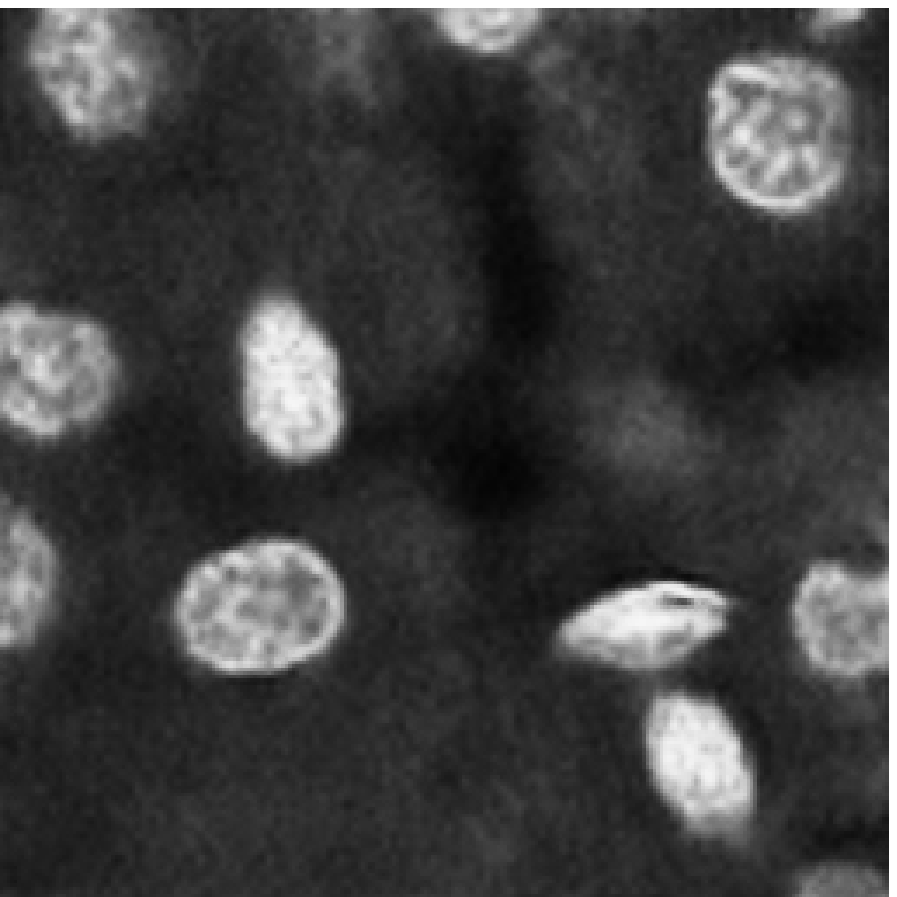,width=2.5cm} %width=3.8cm
}
\subfigure[]
{
   \epsfig{figure=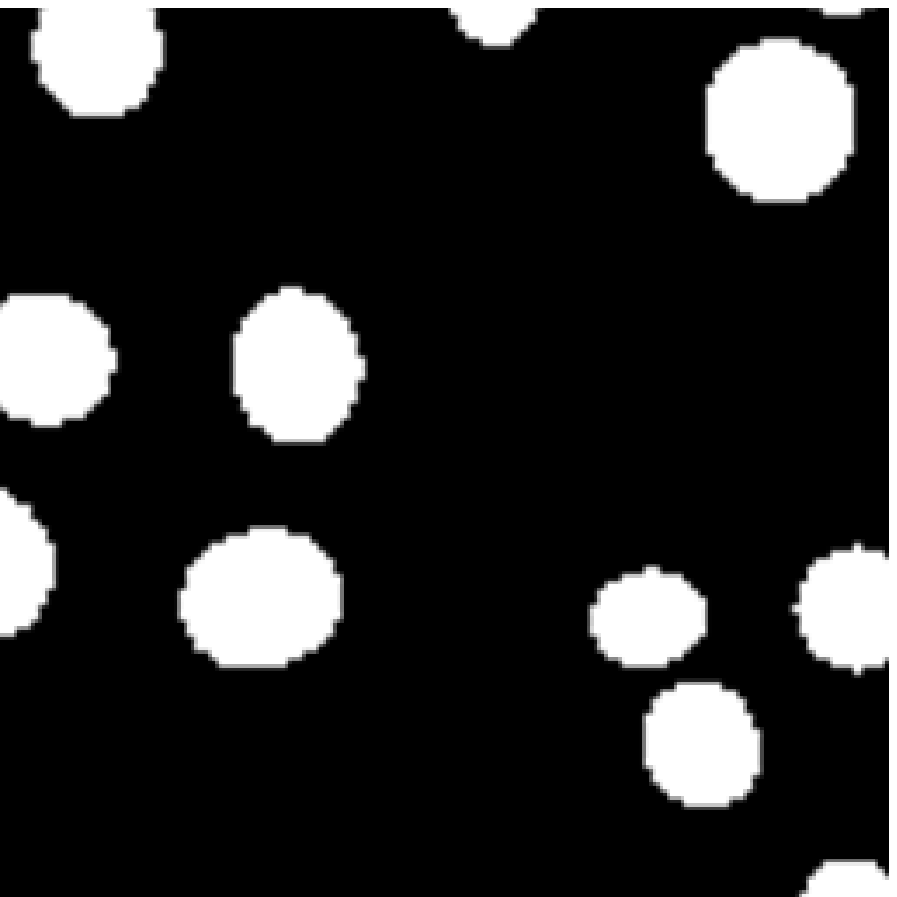,width=2.5cm}
}
%\vspace{-0.05in}
\caption{Slices of the original volume, the synthetic microscopy volume, and the corresponding synthetic binary volume for Data-I and Data-II (a) original image of Data-I, (b) synthetic microscopy image of Data-I, (c) synthetic binary image of Data-I, (d) original image of Data-II, (e) synthetic microscopy image of Data-II, (f) synthetic binary image of Data-II}
\label{fig:synimages}
\vspace{-0.1in}
\end{figure}

% Figure 4
\begin{figure}[htb!]
\centering
\subfigure[]
{
   \epsfig{figure=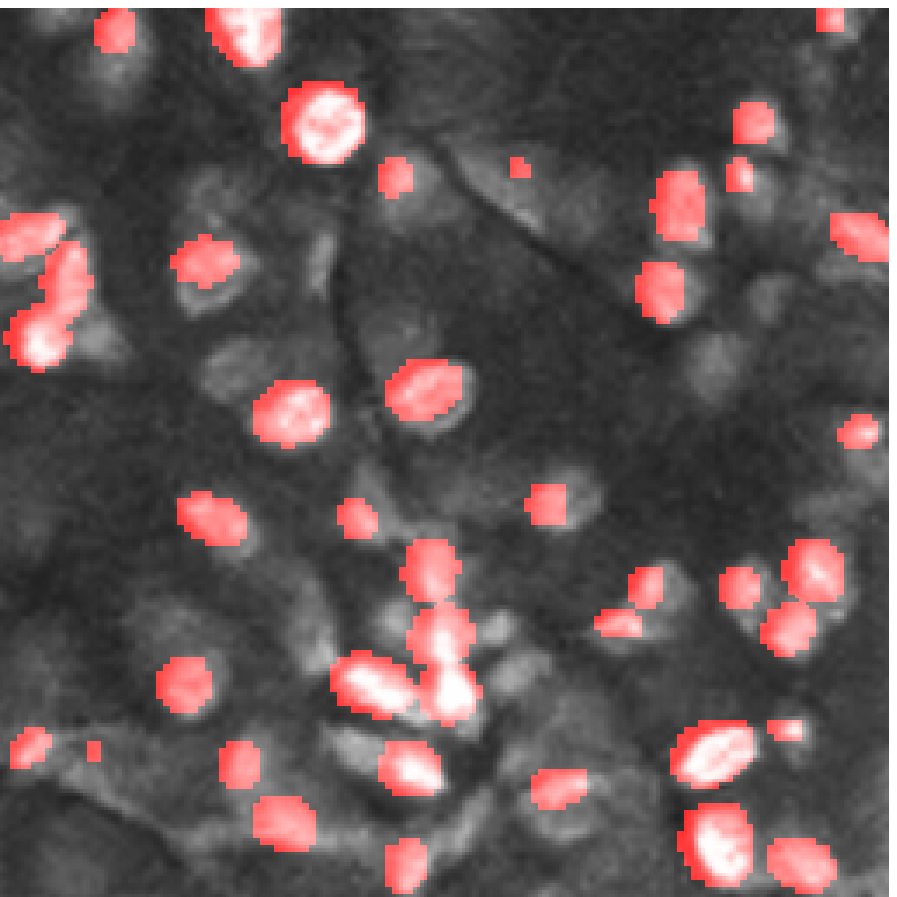,width=3.8cm}
   \label{fig:CycleGANGT}
}
\subfigure[]
{
   \epsfig{figure=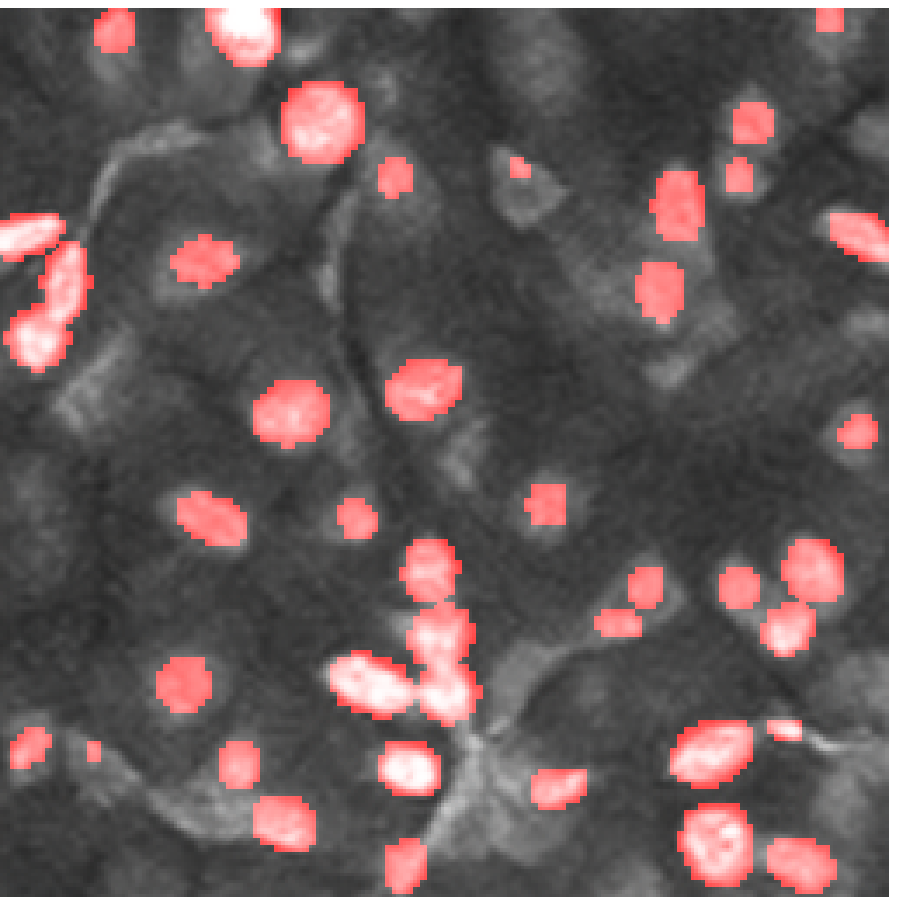,width=3.8cm} %width=3.8cm
   \label{fig:SpCycleGANGT}
}
%\subfigure[]
%{
%   \epsfig{figure=gt.eps,width=2.5cm}
%} 
\caption{A comparison between two synthetic data generation methods overlaid on the corresponding synthetic binary image (a) CycleGAN, (b) SpCycleGAN} 
\label{fig:synimagesGT}
\vspace{-0.1in}
\end{figure}

Figure \ref{fig:synimages} shows the synthetic images generated by our proposed method. The left column indicates original images whereas middle column shows synthetic images artificially generated from corresponding synthetic binary images provided in right column. As can be seen from Figure \ref{fig:synimages}, the synthetic images reflect characteristics of the original microscopy images such as background noise, nuclei shape, orientation and intensity. 

Additionally, two synthetic data generation methods between CycleGAN and SpCycleGAN from the same synthetic binary image are compared in Figure \ref{fig:synimagesGT}. Here, the synthetic binary image is overlaid on the synthetic microscopy image and labeled in red. It is observed that our spatial constraint loss reduces the location shift of nuclei between a synthetic microscopy image and its synthetic binary image. Our realistic synthetic microscopy volumes from SpCycleGAN can be used to train our modified 3D U-Net.

%More importantly, having synthetic binary images, our proposed method is capable of generating corresponding synthetic microscopy images which can be used to train our modified 3D U-Net.

%% Table
\begin{table*}[htb!]
\centering
\caption{Accuracy, Type-I and Type-II errors for known methods and our method on subvolume $1$, subvolume $2$ and subvolume $3$ of Data-I}
\vspace{0.05in}
\renewcommand{\tabcolsep}{4.5pt}
{\begin{tabular}{|c||c|c|c||c|c|c||c|c|c|}
\hline
& \multicolumn{3}{c||}{Subvolume 1} & \multicolumn{3}{c||}{Subvolume 2} & \multicolumn{3}{c|}{Subvolume 3} \\
\hline
{Method} & {Accuracy} & {Type-I} & {Type-II} & {Accuracy} & {Type-I} & {Type-II} & {Accuracy} & {Type-I} & {Type-II} \\
\hline
Method \cite{lorenz2013} & 84.09\% & 15.68\% & 0.23\% & 79.25\% & 20.71\% & 0.04\% & 76.44\% & 23.55\% & 0.01\% \\
\hline
Method \cite{lee2017} & 87.36\% & 12.44\% & 0.20\% & 86.78\% & 13.12\% & 0.10\% & 83.47\% & 16.53\% & 0.00\% \\
\hline
Method \cite{paul2013,rizk2014} & 90.14\% & 9.07\% & 0.79\% & 88.26\% & 11.67\% & 0.07\% & 87.29\% & 12.61\% & 0.10\% \\
\hline
Method \cite{ho2017} & 92.20\% & 5.38\% & 2.42\% & 92.32\% & 6.81\% & 0.87\% & 94.26\% & 5.19\% & 0.55\% \\
\hline
3D Encoder-Decoder & \multirow{3}{*}{93.05\%} & \multirow{3}{*}{3.09\%} & \multirow{3}{*}{3.87\%} & \multirow{3}{*}{91.30\%} & \multirow{3}{*}{5.64\%} & \multirow{3}{*}{3.06\%} & \multirow{3}{*}{94.17\%} & \multirow{3}{*}{3.96\%} & \multirow{3}{*}{1.88\%} \\
+ CycleGAN + BCE &   &   &   &   &   &   &   &   & \\ ($\mu_1 = 0$, $\mu_2 = 1$,$ V = 80$) &   &   &   &   &   &   &   &  & \\
\hline
3D Encoder-Decoder & \multirow{3}{*}{94.78\%} & \multirow{3}{*}{3.42\%} & \multirow{3}{*}{1.79\%} & \multirow{3}{*}{92.45\%} & \multirow{3}{*}{6.62\%} & \multirow{3}{*}{0.92\%} & \multirow{3}{*}{93.57\%} & \multirow{3}{*}{6.10\%} & \multirow{3}{*}{0.33\%} \\
+ SpCycleGAN + BCE &   &   &   &   &   &   &   &   & \\ ($\mu_1 = 0$, $\mu_2 = 1$,$ V = 80$) &   &   &   &   &   &   &   &  & \\
\hline
3D U-Net + SpCycleGAN & \multirow{3}{*}{95.07\%} & \multirow{3}{*}{2.94\%} & \multirow{3}{*}{1.99\%} & \multirow{3}{*}{93.01\%} & \multirow{3}{*}{6.27\%} & \multirow{3}{*}{0.72\%} & \multirow{3}{*}{94.04\%} & \multirow{3}{*}{5.84\%} & \multirow{3}{*}{0.11\%} \\
+ BCE &   &   &   &   &   &   &   &   & \\ ($\mu_1 = 0$, $\mu_2 = 1$,$ V = 80$) &   &   &   &   &   &   &   &  & \\
\hline
3D U-Net + SpCycleGAN & \multirow{3}{*}{94.76\%} & \multirow{3}{*}{3.00\%} & \multirow{3}{*}{2.24\%} & \multirow{3}{*}{93.03\%} & \multirow{3}{*}{6.03\%} & \multirow{3}{*}{0.95\%} & \multirow{3}{*}{94.30\%} & \multirow{3}{*}{5.22\%} & \multirow{3}{*}{0.40\%} \\
+ DICE &   &   &   &   &   &   &   &   & \\ ($\mu_1 = 1$, $\mu_2 = 0$,$ V = 80$) &   &   &   &   &   &   &   &  & \\
\hline
3D U-Net +SpCycleGAN & \multirow{3}{*}{95.44\%} & \multirow{3}{*}{2.79\%} & \multirow{3}{*}{1.76\%} & \multirow{3}{*}{93.63\%} & \multirow{3}{*}{5.73\%} & \multirow{3}{*}{0.64\%} & \multirow{3}{*}{93.90\%} & \multirow{3}{*}{5.92\%} & \multirow{3}{*}{0.18\%} \\
 + DICE and BCE  &   &   &   &   &   &   &   &   & \\  ($\mu_1 = 1$, $\mu_2 = 10$,$ V = 80$) &   &   &   &   &   &   &   &  & \\
\hline
3D U-Net +SpCycleGAN & \multirow{3}{*}{95.37\%} & \multirow{3}{*}{2.77\%} & \multirow{3}{*}{1.86\%} & \multirow{3}{*}{93.63\%} & \multirow{3}{*}{5.69\%} & \multirow{3}{*}{0.68\%} & \multirow{3}{*}{94.37\%} & \multirow{3}{*}{5.27\%} & \multirow{3}{*}{0.36\%} \\
 + DICE and BCE  &   &   &   &   &   &   &   &   & \\  ($\mu_1 = 1$, $\mu_2 = 10$,$ V = 1600$) &   &   &   &   &   &   &   &  & \\
\hline
3D U-Net +SpCycleGAN & \multirow{4}{*}{95.56\%} & \multirow{4}{*}{2.57\%} & \multirow{4}{*}{1.86\%} & \multirow{4}{*}{93.67\%} & \multirow{4}{*}{5.65\%} & \multirow{4}{*}{0.68\%} & \multirow{4}{*}{94.54\%} & \multirow{4}{*}{5.10\%} & \multirow{4}{*}{0.36\%} \\
 + DICE and BCE + PP &   &   &   &   &   &   &   &   & \\  ($\mu_1 = 1$, $\mu_2 = 10$,$ V = 1600$) &   &   &   &   &   &   &   &  & \\ (Proposed method) &   &   &   &   &   &   &   &  & \\
\hline
\end{tabular}
}
\vspace{-0.1in}
\label{tab:accuracy}
\end{table*}

% Figure 5
\begin{figure}[h]
\centering
\subfigure[]
{
   \epsfig{figure=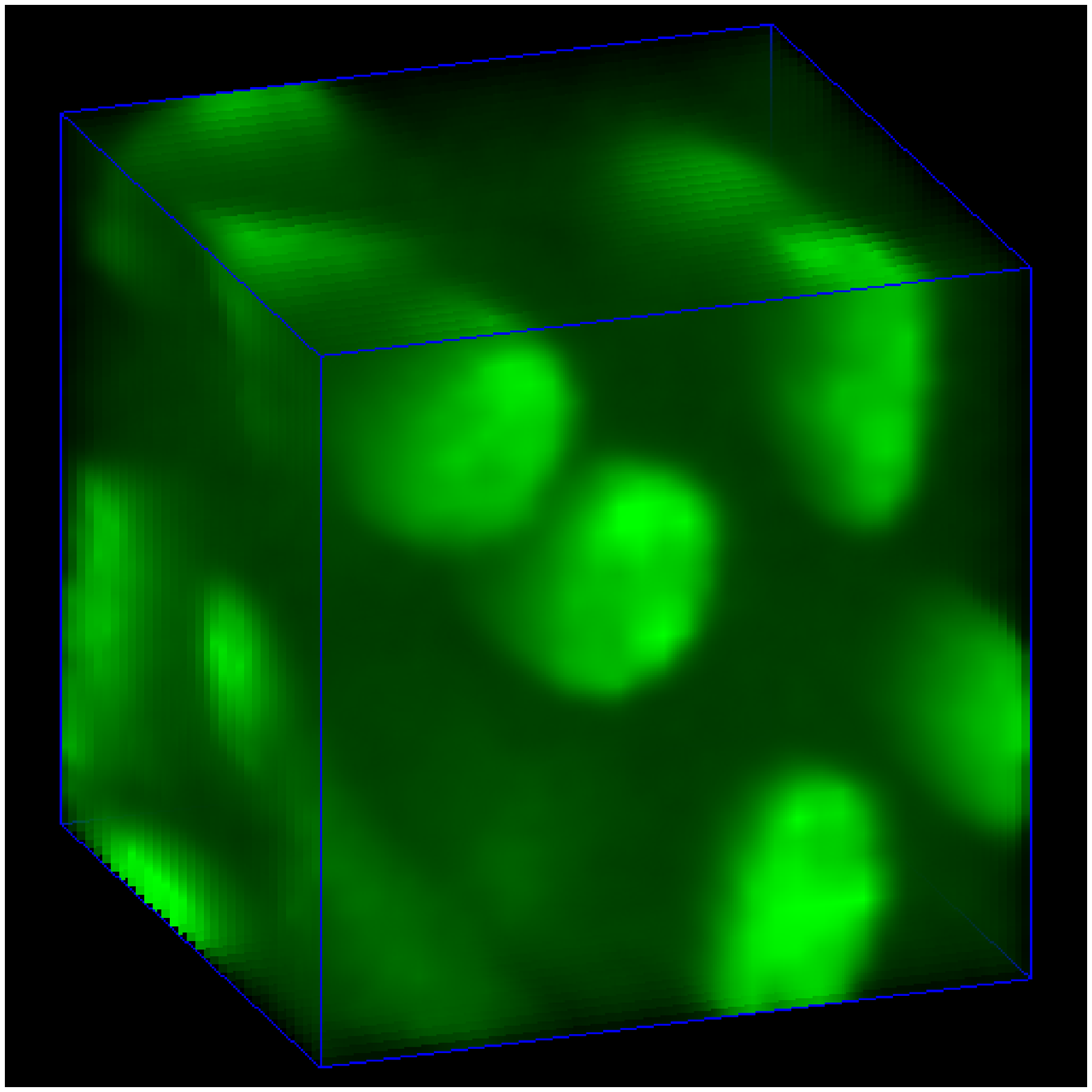,width=3.8cm}
}
\subfigure[]
{
   \epsfig{figure=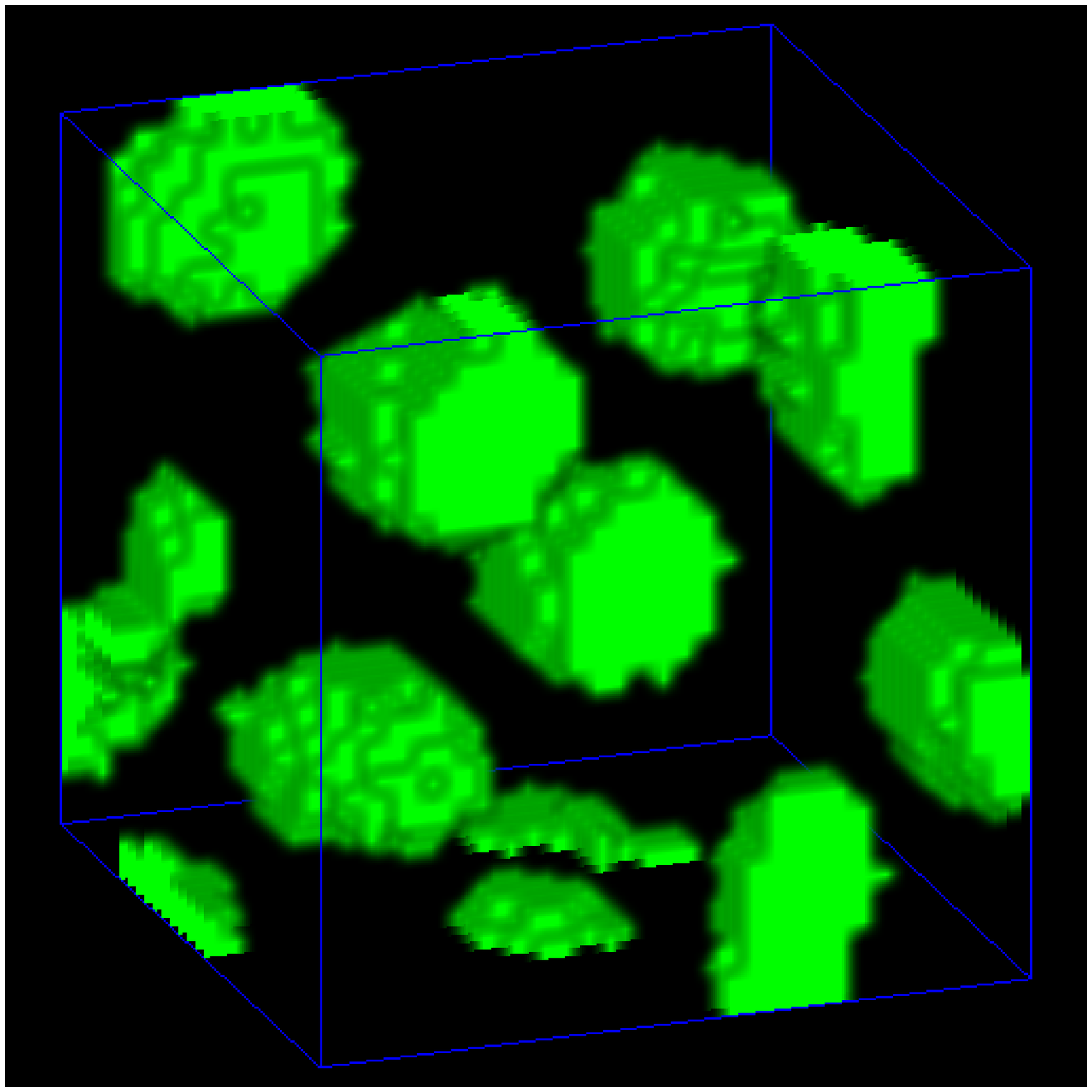,width=3.8cm}
}
\subfigure[]
{
   \epsfig{figure=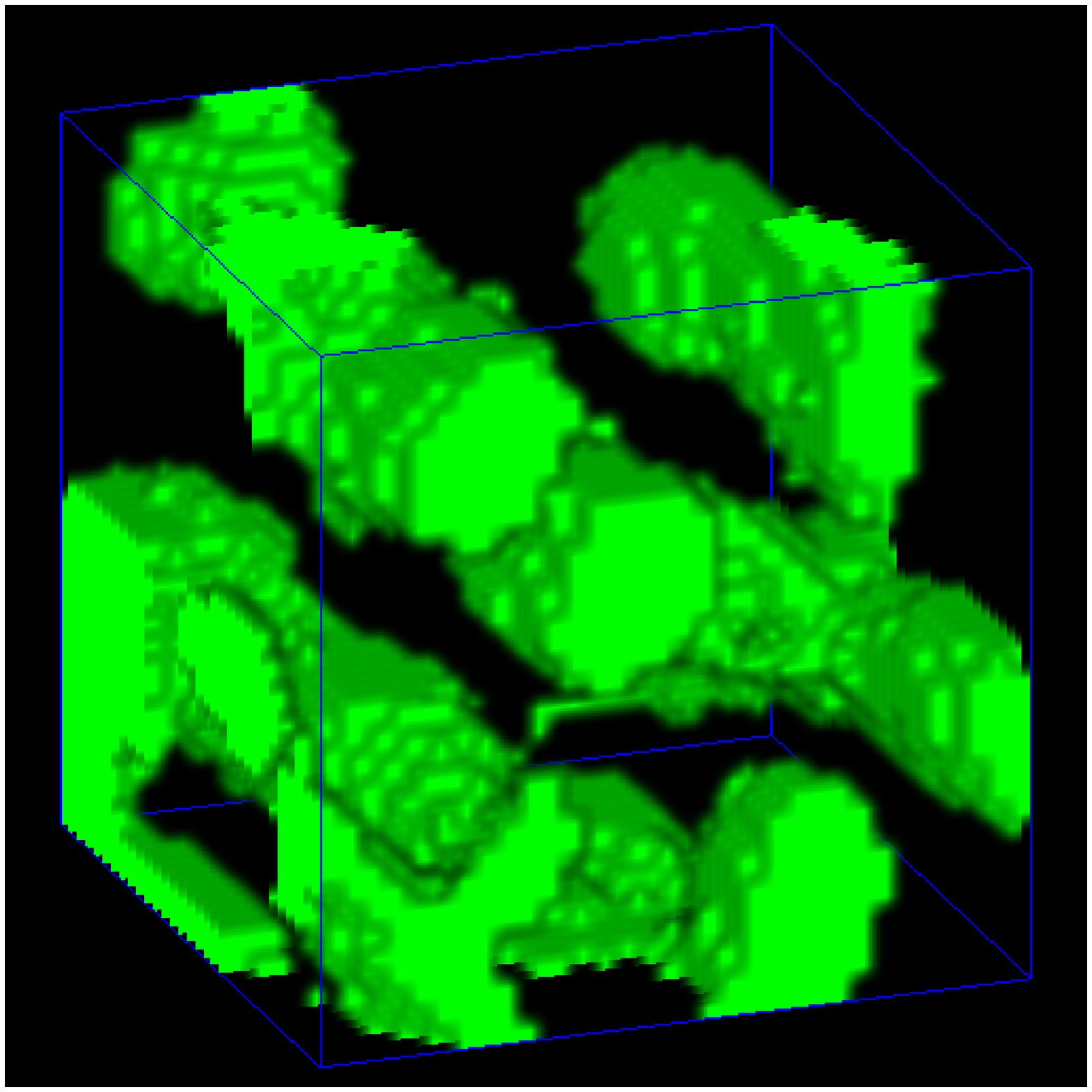,width=3.8cm}
}
\subfigure[]
{
   \epsfig{figure=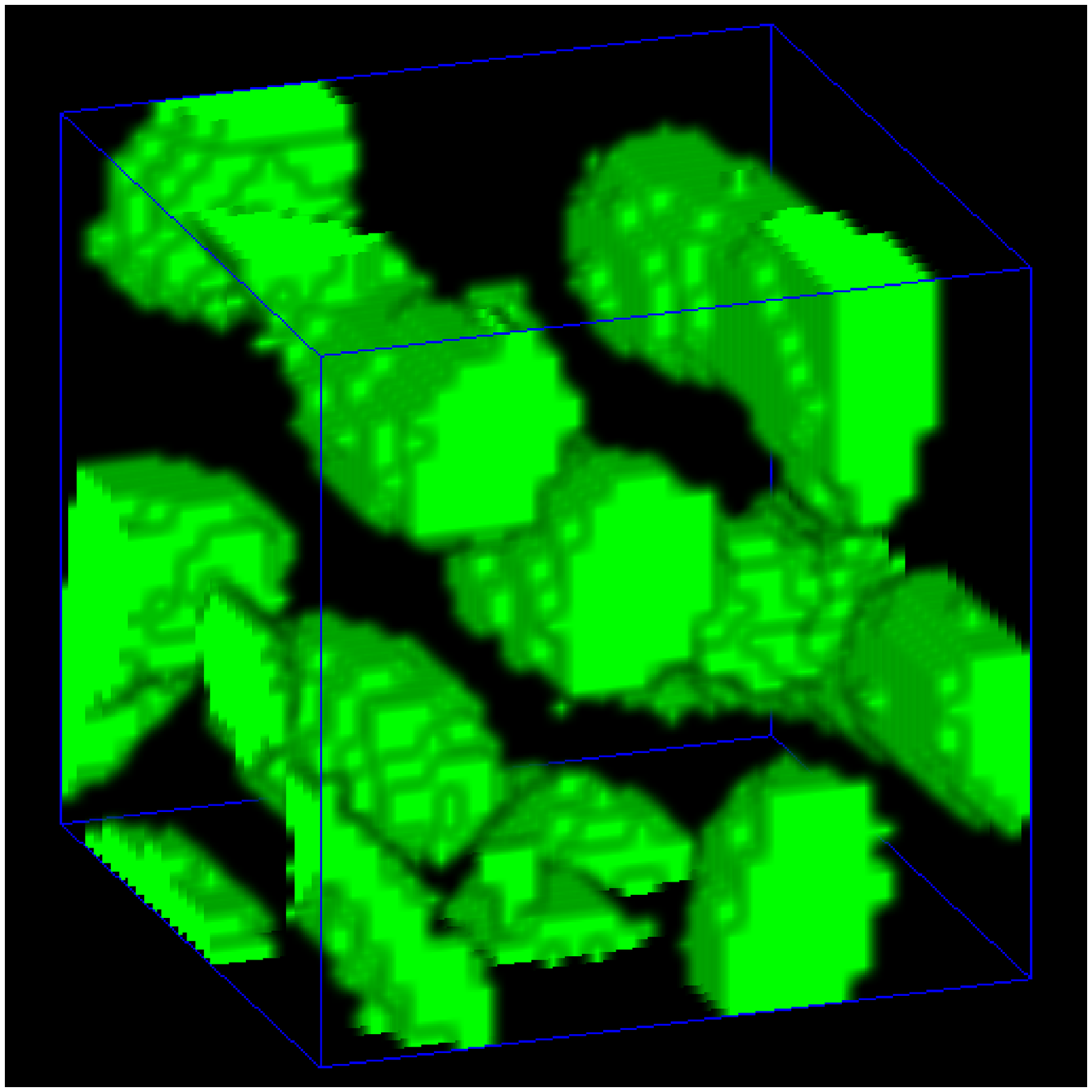,width=3.8cm}
}
\subfigure[]
{
   \epsfig{figure=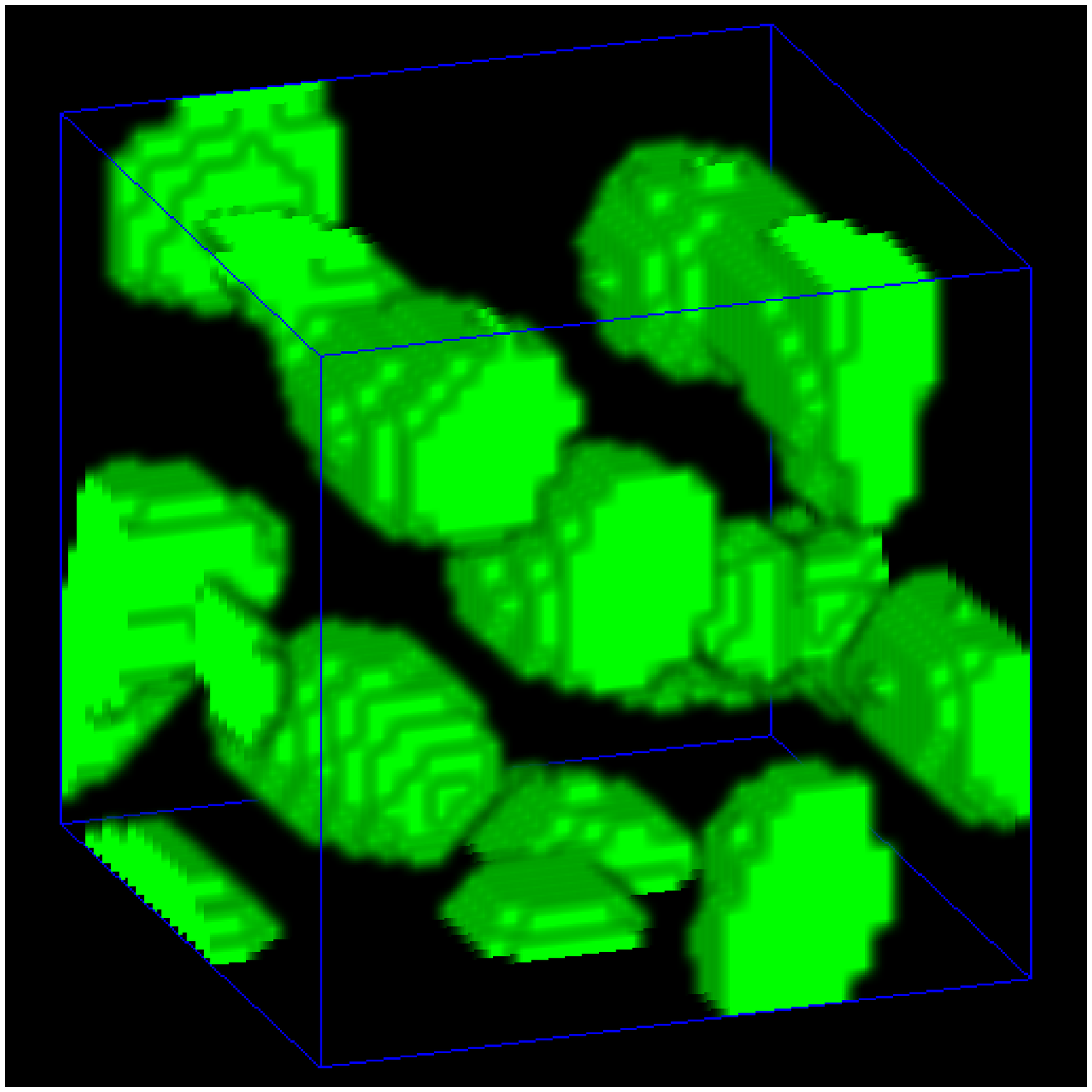,width=3.8cm}
}
\subfigure[]
{
   \epsfig{figure=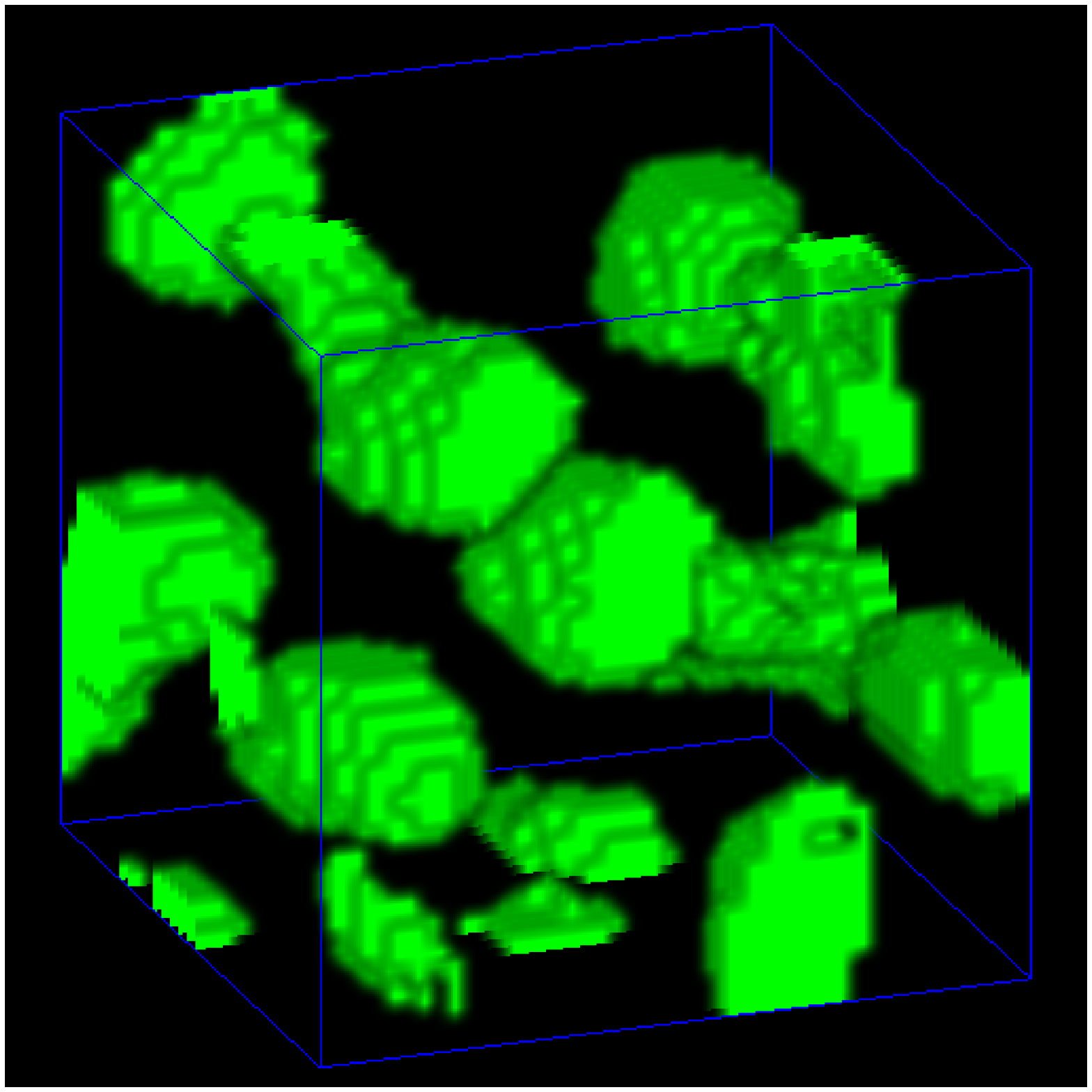,width=3.8cm}
}
\subfigure[]
{
   \epsfig{figure=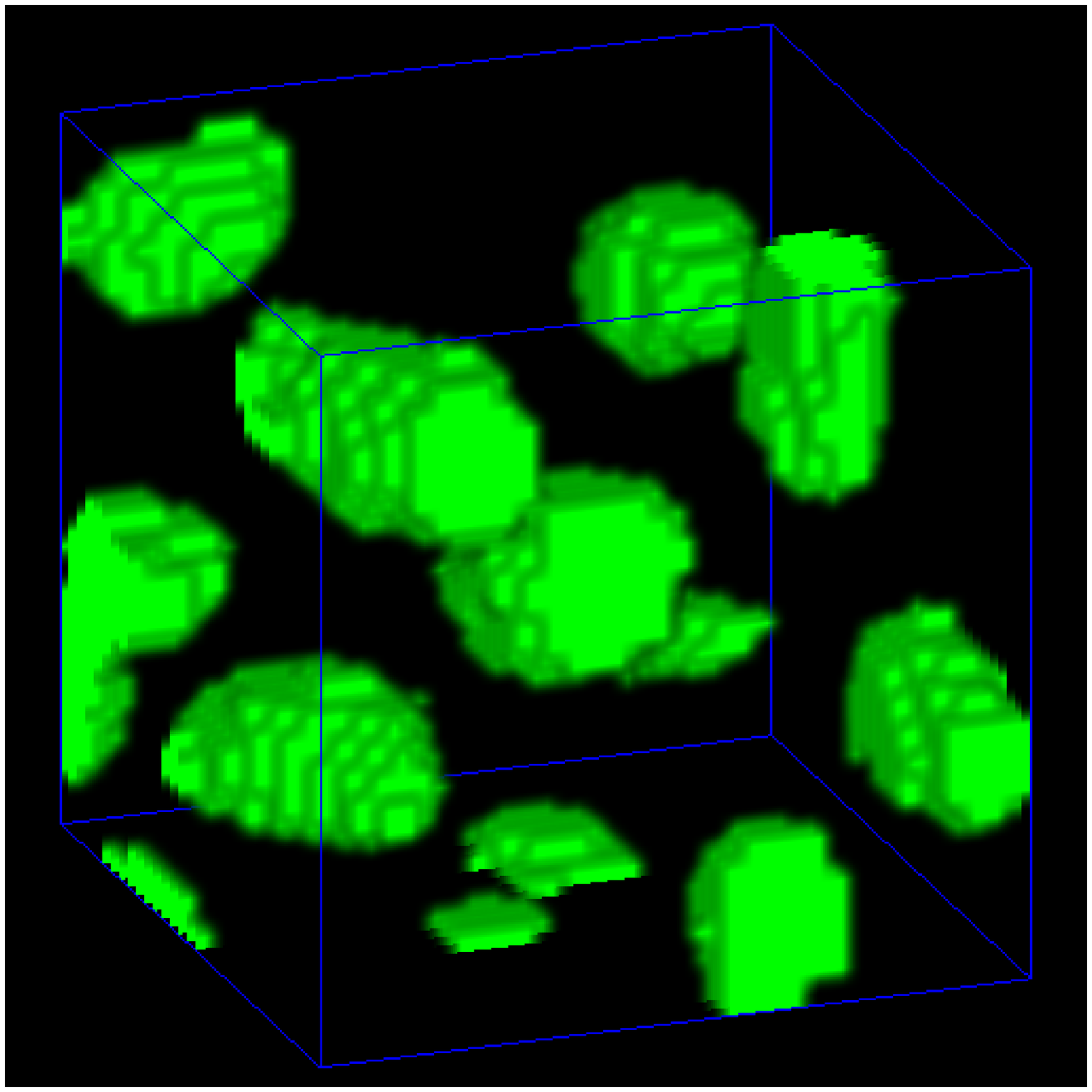,width=3.8cm}
   \label{fig:CycleGAN}
}
\subfigure[]
{
   \epsfig{figure=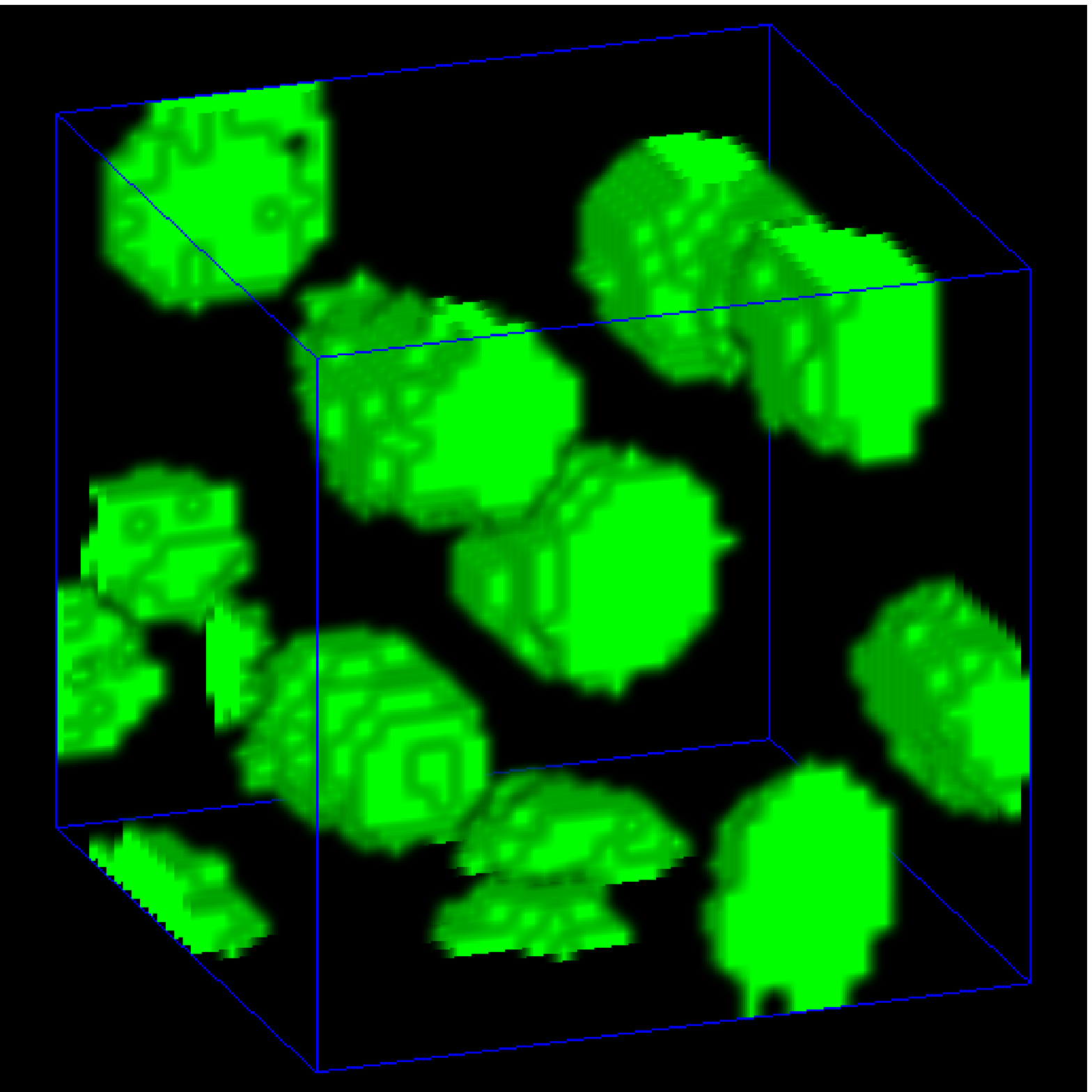,width=3.8cm}
    \label{fig:SpCycleGAN}
}
%\vspace{-0.15in}
\caption{3D visualization of subvolume $1$ of Data-I using Voxx~\cite{clendenon2002} (a) original volume, (b) 3D ground truth volume, (c) 3D active surfaces from \cite{lorenz2013}, (d) 3D active surfaces with inhomogeneity correction from \cite{lee2017}, (e) 3D Squassh from \cite{paul2013, rizk2014}, (f) 3D encoder-decoder architecture from \cite{ho2017}, (g) 3D encoder-decoder architecture with CycleGAN, (h) 3D U-Net architecture with SpCycleGAN (Proposed method)}
\label{fig:comparison1}
\vspace{-0.1in}
\end{figure}

Our proposed method was compared to other 3D segmentation methods including 3D active surface \cite{lorenz2013}, 3D active surface with inhomogeneity correction \cite{lee2017}, 3D Squassh \cite{paul2013,rizk2014}, 3D encoder-decoder architecture \cite{ho2017}, 3D encoder-decoder architecture with CycleGAN. Three original 3D subvolumes of Data-I were selected to evaluate the 
performance of our proposed method. We denote the original volume as subvolume $1$  ($I^{orig}_{\left(241:272,241:272,31:62\right)}$), subvolume $2$ ($I^{orig}_{\left(241:272,241:272,131:162\right)}$), and 
subvolume $3$ ($I^{orig}_{\left(241:272,241:272,231:262\right)}$), respectively. Corresponding groundtruth of each subvolume was hand segmented. Voxx \cite{clendenon2002} was used to visualize the segmentation results in 3D and compared to the manually annotated volumes. In Figure \ref{fig:comparison1}, 3D visualizations of the hand segmented subvolume $1$ and the corresponding segmentation results for various methods were presented. As seen from the 3D visualization in Figure \ref{fig:comparison1}, our proposed method shows the best performance among presented methods visually compared to hand segmented groundtruth volume. 
In general, our proposed method captures only nuclei structure whereas other presented methods falsely detect non-nuclei structures as nuclei. Note that segmentation results in Figure \ref{fig:CycleGAN} yields smaller segmentation mask and suffered from location shift. Our proposed method shown in Figure \ref{fig:SpCycleGAN} outperforms Figure \ref{fig:CycleGAN} since our proposed method uses spatially constrained CycleGAN and takes consideration of the Dice loss and the binary cross-entropy loss.

\begin{figure}[htb!]
\centering
\subfigure[]
{
   \epsfig{figure=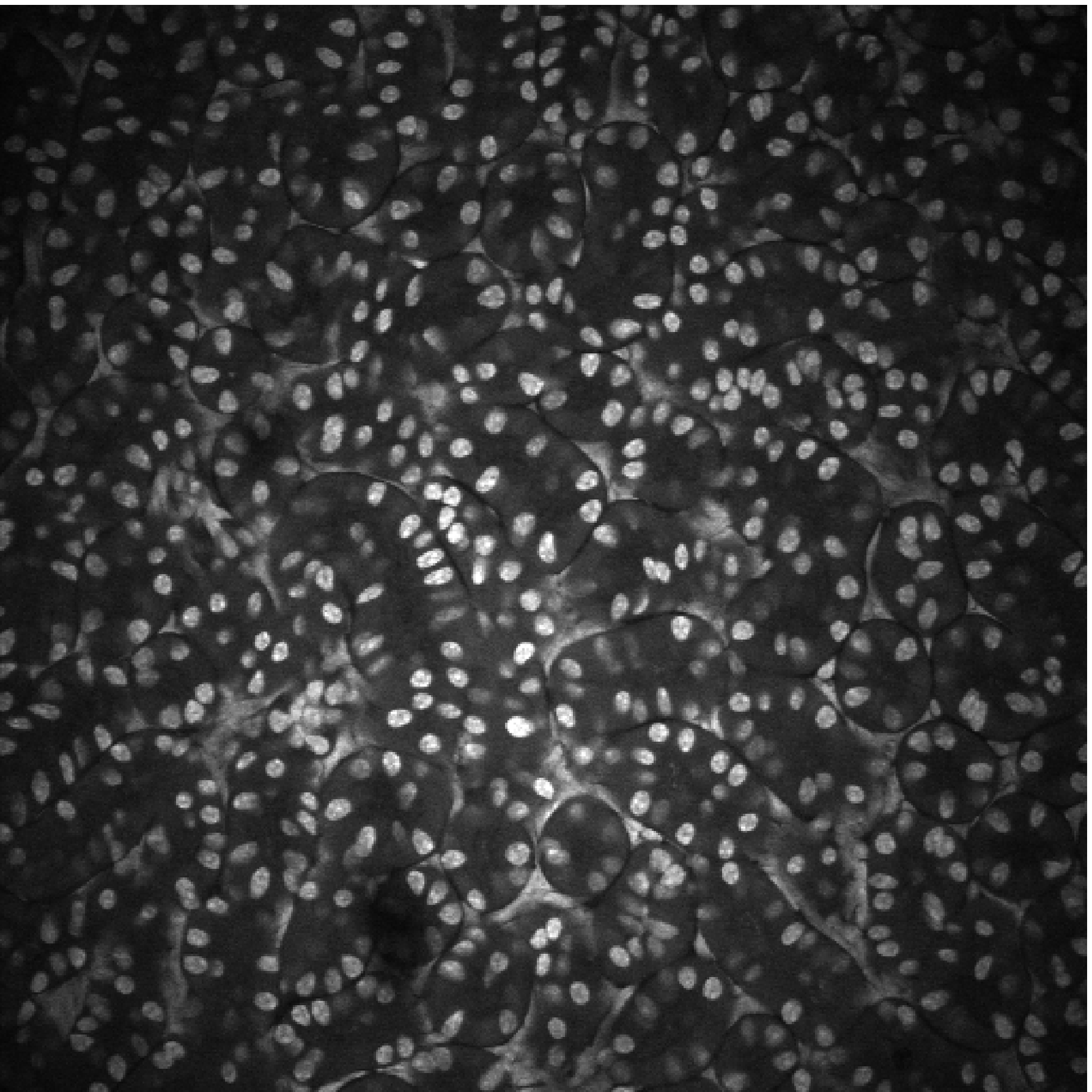,width=3.8cm} %width=3.8cm
}
\subfigure[]
{
   \epsfig{figure=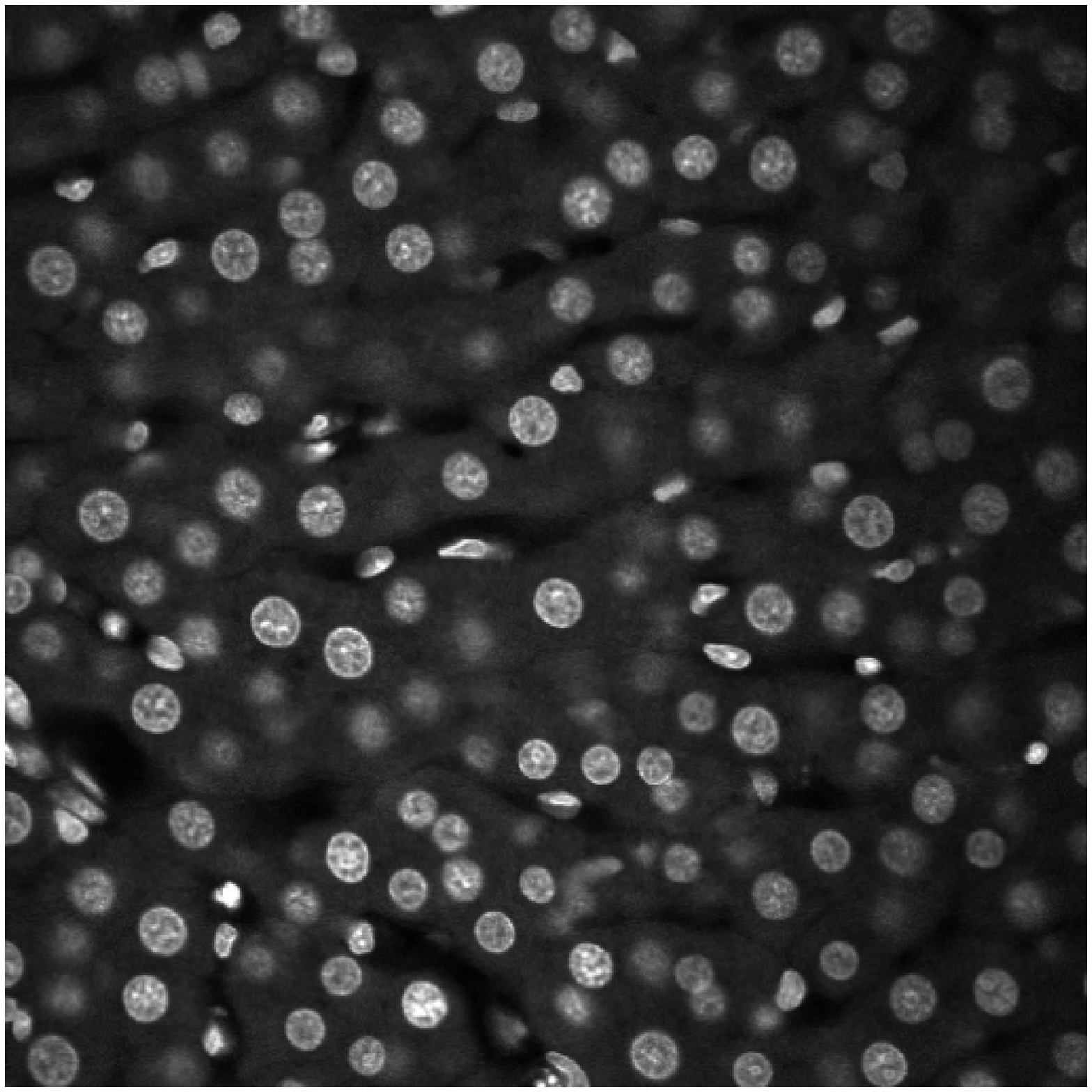,width=3.8cm}
}
\subfigure[]
{
   \epsfig{figure=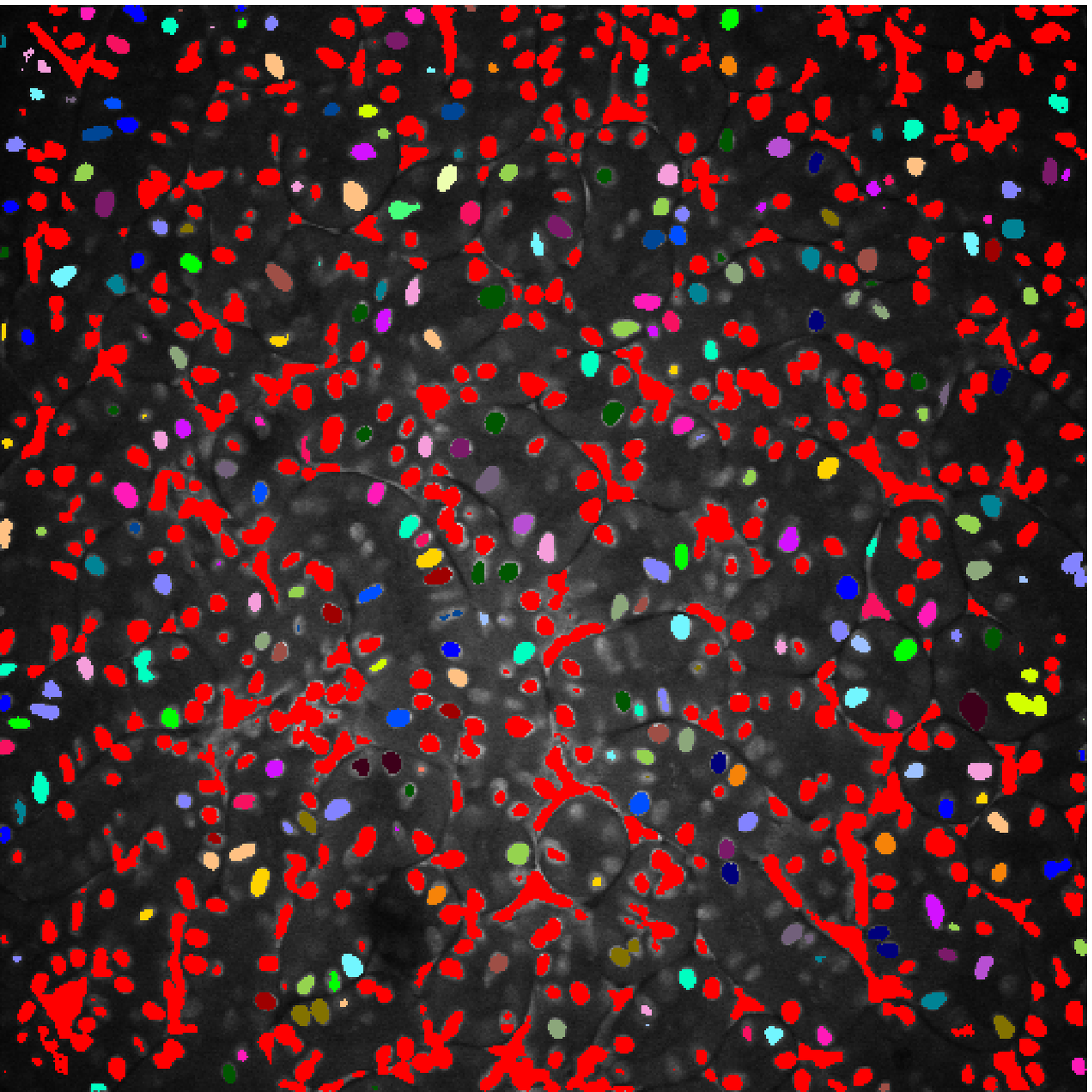,width=3.8cm}
}
\subfigure[]
{
   \epsfig{figure=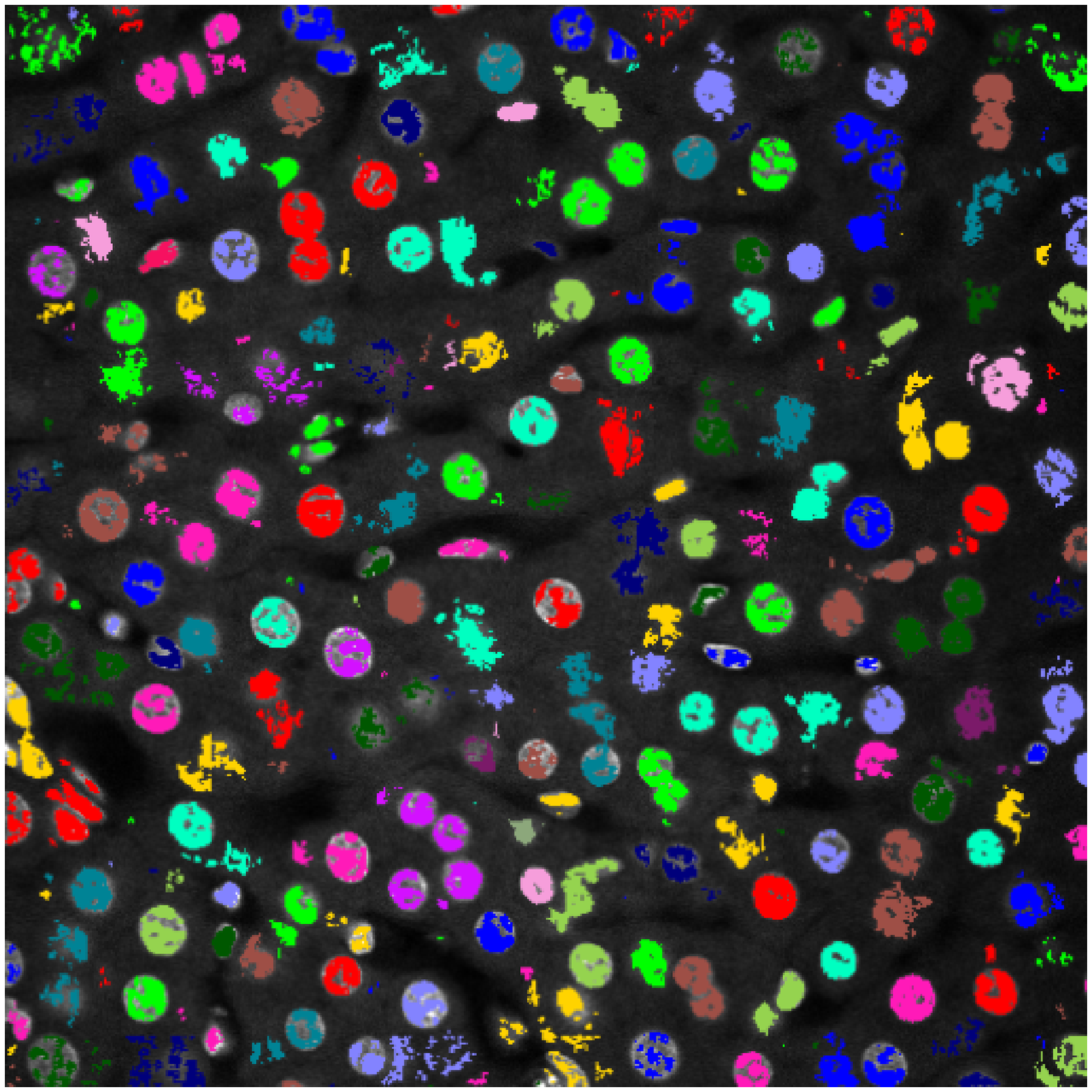,width=3.8cm}
}
\subfigure[]
{
   \epsfig{figure=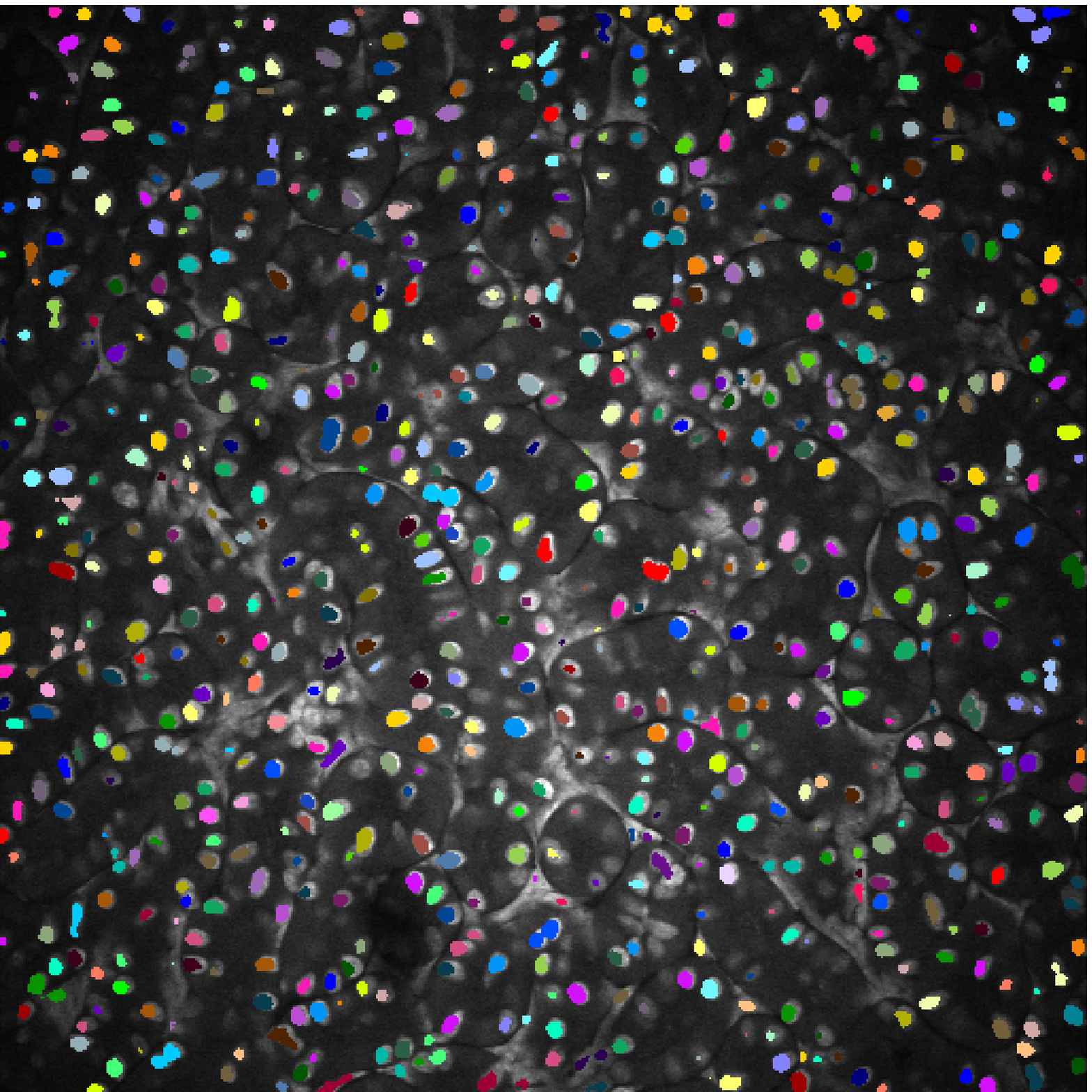,width=3.8cm}
   \label{fig:deep3D+wsm}
}
\subfigure[]
{
   \epsfig{figure=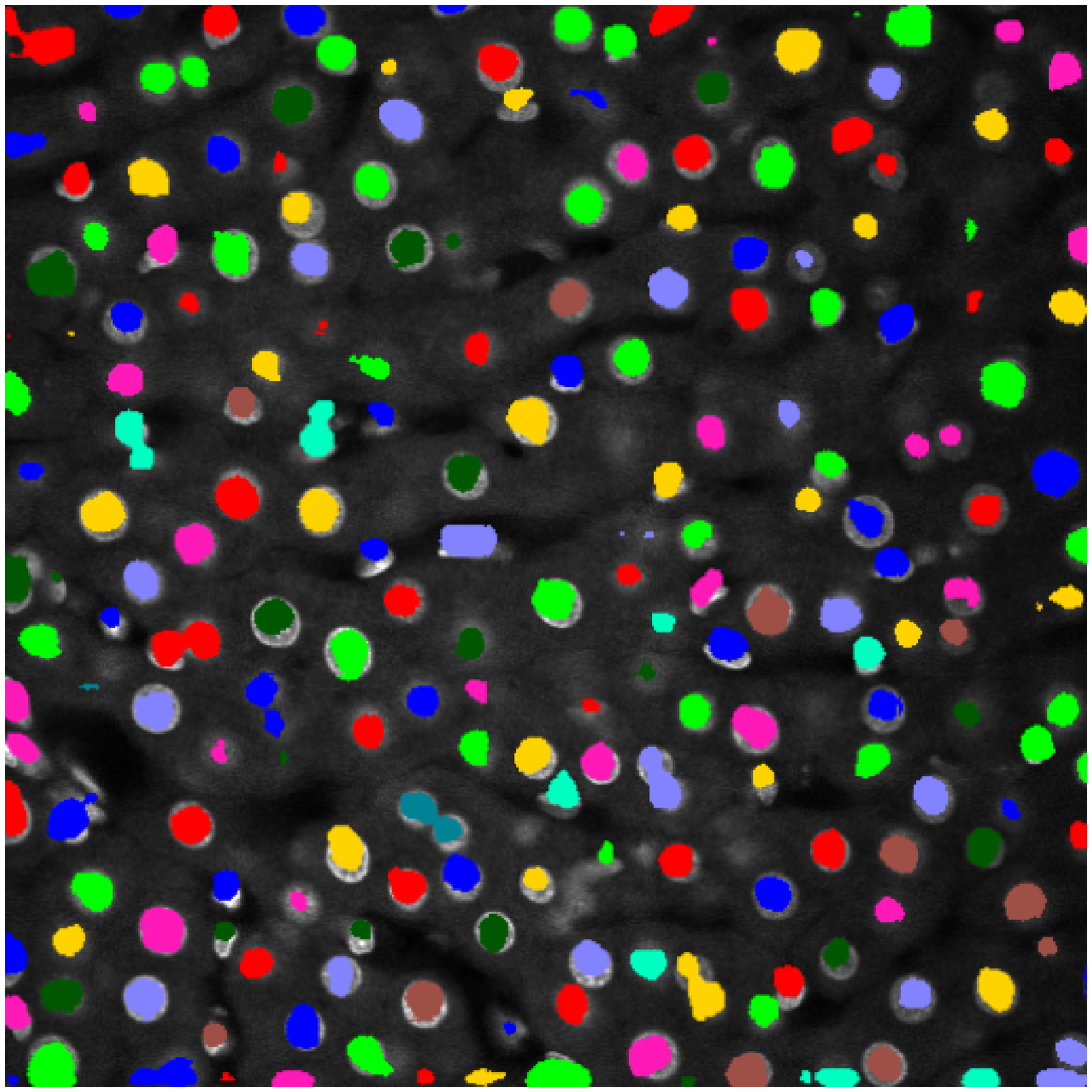,width=3.8cm}
   \label{fig:deep3D+immu}
}
\subfigure[]
{
   \epsfig{figure=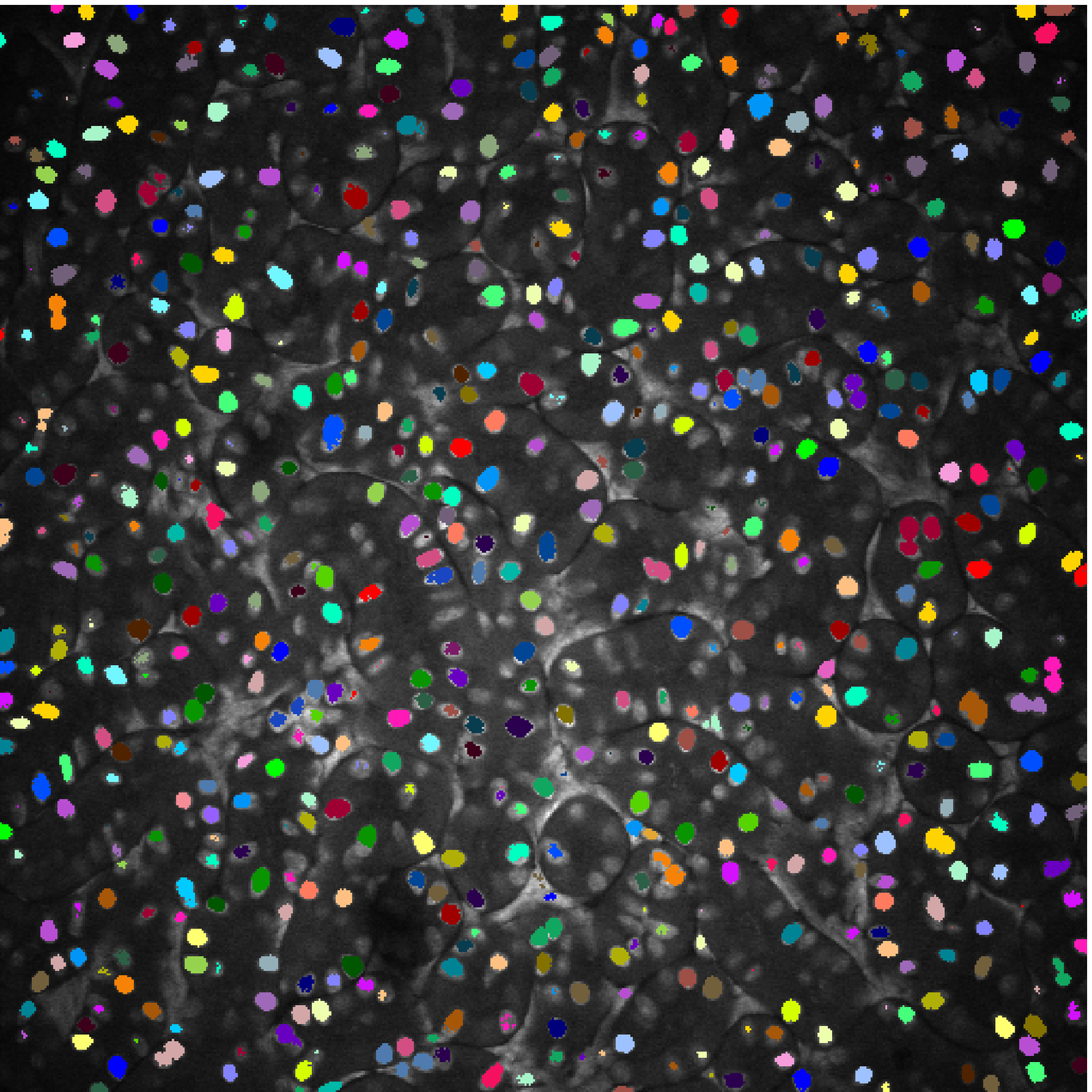,width=3.8cm}
   \label{fig:deep3D++wsm}
}
\subfigure[]
{
   \epsfig{figure=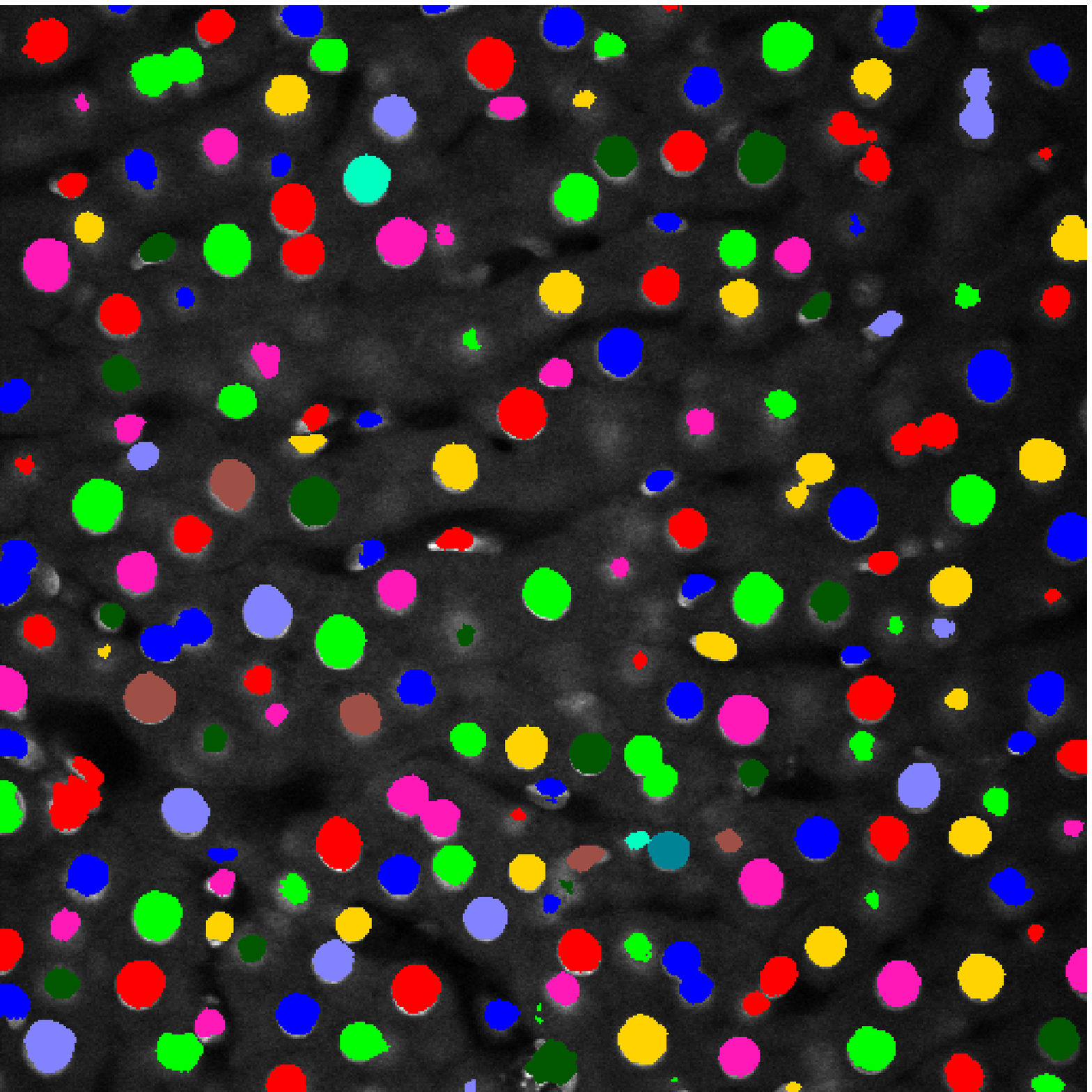,width=3.8cm}
   \label{fig:deep3D++immu}
}
%\vspace{-0.15in}
\caption{Original images and their color coded segmentation results of Data-I and Data-II (a) Data-I $I^{orig}_{z_{66}}$, (b) Data-II $I^{orig}_{z_{31}} $, (c) Data-I $I^{seg}_{z_{66}}$ using \cite{ho2017}, (d) Data-II $I^{seg}_{z_{31}}$ using \cite{ho2017}, (e) Data-I $I^{seg}_{z_{66}}$ using 3D encoder-decoder architecture with CycleGAN, (f) Data-II $I^{seg}_{z_{31}}$ using 3D encoder-decoder architecture with CycleGAN, (g) Data-I $I^{seg}_{z_{66}}$ using 3D U-Net architecture with SpCycleGAN (Proposed method), (h) Data-II $I^{seg}_{z_{31}}$ using 3D U-Net architecture with SpCycleGAN (Proposed method)}
\label{fig:wsm}
\vspace{-0.2in}
\end{figure}

All segmentation results were evaluated quantitatively based on voxel accuracy, Type-I error and Type-II error metrics, using 3D hand segmented volumes. Here, $\text{accuracy} = \frac{n_\text{TP}+n_\text{TN}}{n_\text{total}}$, $\text{Type-I error} = \frac{n_\text{FP}}{n_\text{total}}$, $\text{Type-II error} = \frac{n_\text{FN}}{n_\text{total}}$, where $n_\text{TP}$, $n_\text{TN}$, $n_\text{FP}$, $n_\text{FN}$, $n_\text{total}$ are defined to be the number of true-positives (voxels segmented as nuclei correctly), true-negatives (voxels segmented as background correctly), false-positives (voxels falsely segmented as nuclei), false-negatives (voxels falsely segmented as background), and the total number of voxels in a volume, respectively.

The quantitatively evaluations for the subvolumes are shown in Table \ref{tab:accuracy}. Our proposed method outperforms other compared methods. The smaller Type-I error shows our proposed method successfully rejects non-nuclei structures during segmentation. Also, our proposed method has reasonably low Type-II errors compared to other segmentation methods. Moreover, in this table, we show that our proposed SpCycleGAN creates better paired synthetic volumes which reflects in segmentation accuracy. Instead of 3D encoder-decoder structure, we use 3D U-Net which leads to better results since 3D U-Net has skip connections that can preserve spatial information. In addition, the combination of two loss functions such as the Dice loss and the BCE loss turns out to be better for the segmentation task in our application. In particular, the Dice loss constrains the shape of the nuclei segmentation whereas the BCE loss regulates voxelwise binary prediction. It is observed that training with more synthetic volumes can generalize our method to achieve better segmentation accuracy. Finally, the postprocessing (PP) that eliminates small components helps to improve segmentation performance. 

To make this clear, segmentation results were color coded using 3D connected component labeling and overlaid on the original volumes. The method from \cite{ho2017} cannot distinguish between nuclei and non-nuclei structures including noise. 
This is especially recognizable from segmentation results of Data-I in which multiple nuclei and non-nuclei structures are colored with the same color. As can be observed from Figure \ref{fig:deep3D+wsm} and \ref{fig:deep3D+immu}, segmentation masks are smaller than nuclei size and suffered from location shifts. Conversely, our proposed method shown in Figure \ref{fig:deep3D++wsm} and \ref{fig:deep3D++immu} segments nuclei with the right shape at the correct locations.

%%%%%%%%%%%%%%%%%%%%%%%%%%%%%%%%%%%%%%%%%%%%%%%%%%%%%%%%
%\vspace{-0.05in}
\section{Conclusion}
\label{sec:conclusion}
%\vspace{-0.05in}

In this paper we presented a modified 3D U-Net nuclei segmentation method using paired synthetic volumes. The training was done using synthetic volumes generated from a spatially constrained CycleGAN. The combination of the Dice loss and the binary cross-entropy loss functions are optimized during training. We compared our proposed method to various segmentation methods and with manually annotated 3D groundtruth from real data. The experimental results indicate that our method can successfully distinguish between non-nuclei and nuclei structure and capture nuclei regions well from various microscopy volumes. One drawback of our proposed segmentation method is that our method cannot separate nuclei if they are physically touching to each other. In the future, we plan to develop nuclei localization method to identify overlapping nuclei to individuals.

%%%%%%%%%%%%%%%%%%%%%%%%%%%%%%%%%%%%%%%%%%%%%%%%%%%%%%%%
\section{Acknowledgments}
This work was partially supported by a George M. O’Brien Award from the National Institutes of Health under grant NIH/NIDDK P30 DK079312 and the endowment of the Charles William Harrison Distinguished Professorship at Purdue University.

Data-I was provided by Malgorzata Kamocka of Indiana University and was collected at the Indiana Center for Biological
Microscopy. 

Address all correspondence to Edward J. Delp, ace@ecn.purdue.edu

%%%%%%%%%%%%%%%%%%%%%%%%%%%%%%%%%%%%%%%%%%%%%%%%%%%%%%%%
%\clearpage
%\vspace{-0.1in}
%\bibliographystyle{ieee}
%\bibliography{refs}

{\small
\bibliographystyle{ieee}
\bibliography{ref}
}

\end{document}